%% file: main.tex
\documentclass[10pt,twocolumn,letterpaper]{article}

\usepackage[pagenumbers]{cvpr} 

\usepackage{graphicx}
\usepackage{amsmath}
\usepackage{amssymb}
\usepackage{booktabs}
\usepackage[dvipsnames]{xcolor}
\usepackage[accsupp]{axessibility}
\definecolor{colortab0}{HTML}{1F77B4}
\definecolor{colortab1}{HTML}{FF7F0E}
\definecolor{colortab2}{HTML}{2CA02C}
\definecolor{colortab3}{HTML}{D62728}

\usepackage[pagebackref,breaklinks,colorlinks]{hyperref}

\usepackage[capitalize]{cleveref}
\crefname{section}{Sec.}{Secs.}
\Crefname{section}{Section}{Sections}
\Crefname{table}{Table}{Tables}
\crefname{table}{Tab.}{Tabs.}

\def\cvprPaperID{8910} 
\def\confName{CVPR}
\def\confYear{2022}

\input{preamble}
\begin{document}
 \input{sec/0_metadata}
 \maketitle
 \input{sec/0_abstract}%
 \input{sec/1_introduction}
\input{sec/2_related} \input{sec/3_method}
\input{sec/4_results}%
 \input{sec/limitations}%
 \input{sec/5_conclusions}
 {
     \clearpage
     \small
     \bibliographystyle{ieeetr_fullname}
     \bibliography{macros,main,supplemental}
 }
 \input{sec/X_supplementary}

\end{document}

%% file: preamble.tex
\usepackage{overpic}
\usepackage{enumitem} 
\usepackage{overpic} 
\usepackage{color}

\usepackage{array}
\usepackage{adjustbox}
\usepackage{stfloats}
\usepackage{array}
\usepackage{tabularx}
\usepackage{float}
\usepackage{longtable}

\definecolor{turquoise}{cmyk}{0.65,0,0.1,0.3}
\definecolor{purple}{rgb}{0.65,0,0.65}
\definecolor{dark_green}{rgb}{0, 0.5, 0}
\definecolor{orange}{rgb}{0.8, 0.6, 0.2}
\definecolor{red}{rgb}{0.8, 0.2, 0.2}
\definecolor{darkred}{rgb}{0.6, 0.1, 0.05}
\definecolor{blueish}{rgb}{0.0, 0.3, .6}
\definecolor{light_gray}{rgb}{0.7, 0.7, .7}
\definecolor{pink}{rgb}{1, 0, 1}
\definecolor{greyblue}{rgb}{0.25, 0.25, 1}

\usepackage{blindtext}

\renewcommand{\paragraph}[1]{\vspace{1em}\noindent\textbf{#1}.}

%% file: sec/0_metadata.tex
\title{CNN Filter DB: An Empirical Investigation of Trained Convolutional Filters}

\author{Paul Gavrikov$^1$%
\thanks{Funded by the Ministry for Science, Research and Arts, Baden-Wuerttemberg, Grant 32-7545.20/45/1 (Q-AMeLiA).
\newline The authors also thank Margret Keuper for her support and encouragement to submit this work.}
\ and Janis Keuper$^{1,2,3}$\footnotemark[1]\\
$^1$IMLA, Offenburg University, $^2$CC-HPC, Fraunhofer ITWM, $^3$Fraunhofer Research Center ML\\
{\tt\small \{first.last\}@hs-offenburg.de}
}

%% file: sec/0_abstract.tex
\begin{abstract}
\noindent Currently, many theoretical as well as practically relevant questions towards the transferability and robustness of Convolutional Neural Networks (CNNs) remain unsolved. While ongoing research efforts are engaging these problems from various angles, in most computer vision related cases these approaches can be generalized to investigations of the effects of distribution shifts in image data.\\  
In this context, we propose to study the shifts in the learned weights of trained CNN models. Here we focus on the properties of the distributions of dominantly used $3\times 3$ convolution filter kernels. We collected and publicly provide a dataset with over 1.4 billion filters from hundreds of trained CNNs, using a wide range of datasets, architectures, and vision tasks. 
In a first use case of the proposed dataset, we can show highly relevant properties of many publicly available pre-trained models for practical applications: I) We analyze distribution shifts (or the lack thereof) between trained filters along different axes of meta-parameters, like visual category of the dataset, task, architecture, or layer depth. Based on these results, we conclude that model pre-training can succeed on arbitrary datasets if they meet size and variance conditions. II) We show that many pre-trained models contain degenerated filters which make them less robust and less suitable for fine-tuning on target applications. 

{\noindent\normalfont\textbf{Data \& Project website:} \url{https://github.com/paulgavrikov/cnn-filter-db}}
 
\end{abstract}

%% file: sec/1_introduction.tex
\section{Introduction}
\label{sec:intro}

\noindent Despite their overwhelming success in the application to various vision tasks, the practical deployment of convolutional neural networks (CNNs) is still suffering from several inherent drawbacks. Two prominent examples are I) the dependence on very large amounts of annotated training data \cite{sun2017revisiting}, which is not available for all target domains and is expensive to generate; and II) still widely unsolved problems with the robustness and generalization abilities of CNNs \cite{akhtar2018threat} towards shifts of the input data distributions. One can argue that both problems are strongly related, since a common practical solution to I) is the fine-tuning \cite{tajbakhsh2016convolutional} of pre-trained models by small datasets from the actual target domain. This results in the challenge to find suitable pre-trained models based on data distributions that are ``as close as possible'' to the target distributions. Hence, both cases (I+II) imply the need to model and observe distribution shifts in the contexts of CNNs.\\
In this paper, we propose not to investigate these shifts in the input (image) domain, but rather in the 2D filter-kernel distributions of the CNNs themselves. We argue that e.g. the distributions of trained convolutional filters in a CNN, which implicitly reflect the sub-distributions of the input image data, are more suitable and easier accessible representations for this task.
\begin{figure}
  \centering
  \includegraphics[width=0.99\columnwidth]{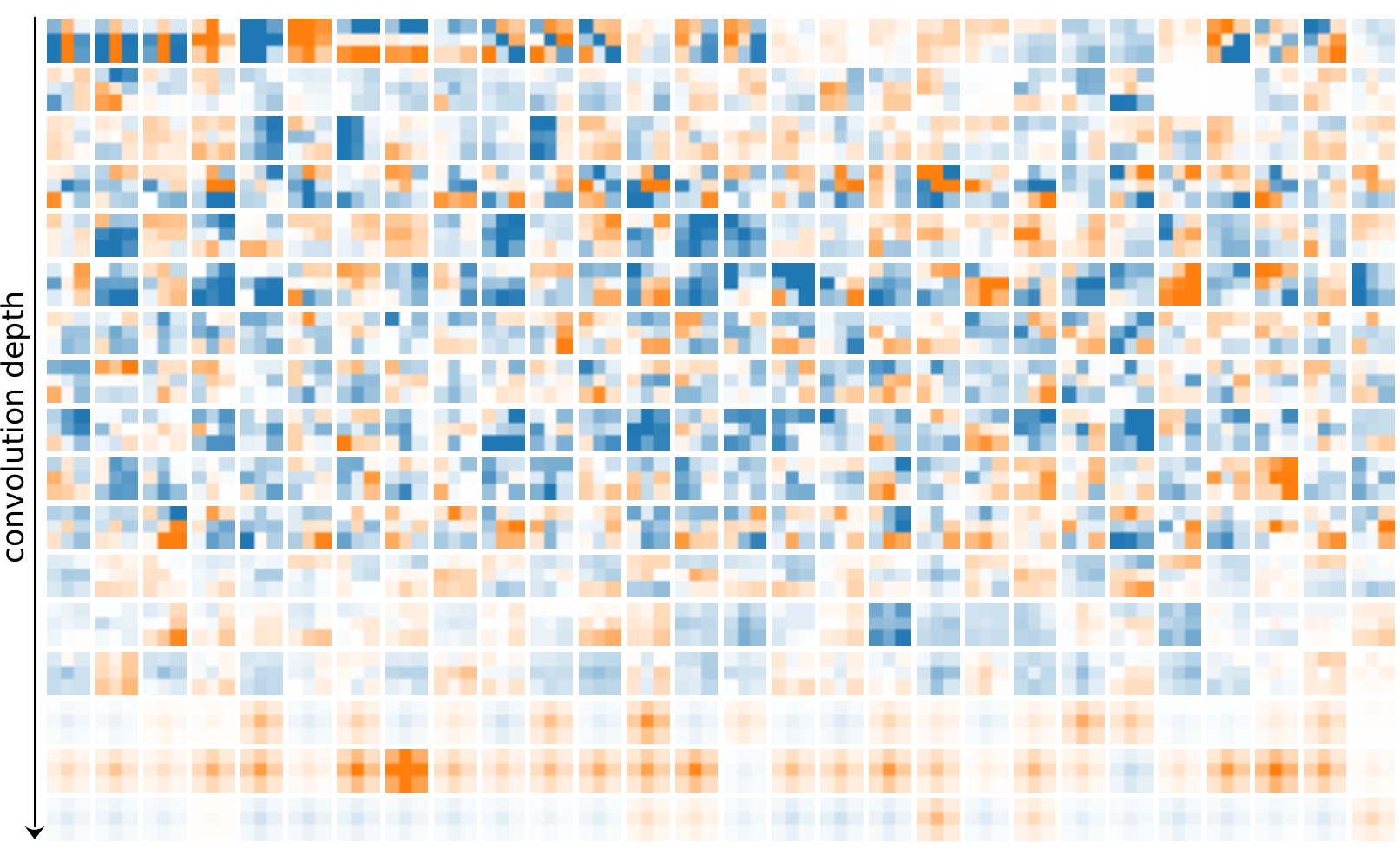}
  \caption{First $3\times 3$ filters extracted of each convolution layer in a \textit{ResNet-18} trained on \textit{CIFAR-10}. The filters show a clear loss of diversity and increasing sparsity with depth. The colormap range is determined layer-wise by the absolute peak weight of all filters in that layer.}
  \label{fig:teaser}
\end{figure}
In order to foster systematic investigations of learned filters, we collected and publicly provide a dataset of over 1.4  billion  filters  with meta data from  hundreds  of trained  CNNs,  using  a  wide  range  of  data  sets,  architectures, and vision tasks. To show the scientific value of this new data source, we conduct a first analysis and report a series of novel insights into widely used CNN models. Based on our presented methods we show that many publicly provided models suffer from degeneration. We show that overparameterization leads to sparse and/or non-diverse filters (\cref{fig:teaser}), while robust training increases filter diversity, and reduces sparsity. Our results also show that learned filters do not significantly differ across models trained for various tasks, except for extreme outliers such as GAN-Discriminators. Models trained on datasets of different visual categories do not significantly drift either. Most shifts in studied models are due to degeneration, rather than an actual difference in structure. Therefore, our results imply that pre-training can be performed independent of the actual target data, and only the amount of training data and its diversity matters. This is inline with recent findings that models can be pre-trained even with images of fractals \cite{KataokaACCV2020}.
For classification models we show that the most variance in learned filters is found in the beginning and end of the model, while object/face detection models only show significant variance in early layers. Also, the most specialized filters are found in the last layers. We summarize our key  contributions as follows:
\begin{itemize}[leftmargin=*]
\setlength\itemsep{-.3em}
\item Publication of a diverse database of over 1.4B $3\times 3$ convolution filters alongside with relevant meta information of the extracted filters and models \cite{zenodo_data}.
\item Presentation of a data-agnostic method based on sparsity and entropy of filters to find ``degenerated'' convolution layers due to overparameterization or non-convergence of trained CNN models.
\item Showing that publicly available models often contain degenerated layers and can therefore be questionable candidates for transfer tasks.
\item Analysis of distribution shifts in filters over various groups, providing insights that formed filters are fairly similar across a wide-range of examined groups.
\item Showing that the model-to-model shifts that exist in classification models are, contrary to the predominant opinion, not only seen in deeper layers but also in the first layers.
\end{itemize}

\paragraph{Paper organization}
We give an overview of our dataset and its collection process in \cref{sec:method},  followed by an introduction of methods studying filter structure, distributions shifts, and layer degeneration such as randomness, low variance in filter structure, and high sparsity of filters. 
Then in \cref{sec:results} we apply these methods to our collected data. We show the impact of overparameterization and robust training on filter degeneration and provide intuitions for threshold finding. Then we analyze filter structures by determining a suitable filter basis and looking into reproducibility of filters in training, filter formation during training, and an analysis of distribution shifts for various dimensions of the collected meta-data.
We discuss limitations of our approach in \cref{sec:limitations} and, finally, draw conclusions in \cref{sec:conclusions}.

%% file: sec/2_related.tex
\section{Related Work}
\label{sec:related}
\noindent We are unaware of any systematic, large scale analysis of learned filters across a wide range of datasets, architectures and task such as the one performed in this paper. However, there are of course several partially overlapping aspects of our analysis that have been covered in related works:\\
\noindent\textbf{Filter analysis.}
An extensive analysis of features, connections, and their organization extracted from trained \textit{InceptionV1} \cite{inception} models was presented in \cite{Olah2020,olah2020an,cammarata2020curve,olah2020naturally,schubert2021high-low,cammarata2021curve,voss2021visualizing,voss2021branch,petrov2021weight}. The authors claim different CNNs will form similar features and circuits even when trained for different tasks.\\
\noindent\textbf{Transfer learning.}
A survey on transfer learning for image classification CNNs can be found in \cite{hussain2019} and general surveys for other tasks and domains are available in \cite{5288526,zhuang2020comprehensive}. The authors of \cite{Yosinski2014} studied learned filter representations in \textit{ImageNet1k} classification models and presented the first approaches towards transfer learning. They argued that different CNNs will form similar filters in early layers which will mostly resemble gabor-filters and color-blobs, while deeper layers will capture specifics of the dataset by forming increasingly specialized filters.
\cite{Aygun_2017_ICCV} captured convolution filter pattern distributions with Gaussian Mixture Models to achieve cross-architecture transfer learning. \cite{tayyab2019basisconv} demonstrated that convolutions filters can be replaced by a fixed filter basis that $1\times 1$ convolution layers blend.\\
\noindent\textbf{Pruning criteria.}
Although we do not attempt pruning, our work overlaps with pruning techniques as they commonly rely on estimation criteria to understand which parameters to compress. These either rely on data-driven computation of a forward-pass \cite{alain2018understanding, luo2017thinet, 8237417,ijcai2018-336,8953628}, or backward-propagation \cite{8579056, 8954199}, or estimate importance solely based on the numerical weight (typically any $\ell$-norm) of the parameters \cite{8953939,han2015learning,li2017pruning,8953212,ijcai2018-309}.\\
\noindent\textbf{CNN distribution shifts.} A benchmark for distribution shifts that arise in real-world applications is provided in \cite{pmlr-v139-koh21a} and \cite{taori2020measuring} measured robustness to natural distribution shifts of 204 \textit{ImageNet1k} models. The authors concluded that robustness to real-world shifts is low.
Lastly, \cite{Djolonga_2021_CVPR} studied the correlation between transfer performance and distribution shifts of image classification models and find that increasing training set and model capacity increases robustness to distribution shifts.

%% file: sec/3_method.tex
\section{Methods}
\label{sec:method}

\subsection{Collecting filters}
\noindent We collected a total of \textbf{647 publicly available CNN models} from \cite{rw2019timm, croce2021robustbench, NEURIPS2019_9015} and other sources that have been pre-trained for various 2D visual tasks\footnote{\label{supp}For more details refer to the supplementary materials.}. In order to provide a heterogeneous and diverse representation of convolution filters ``in the wild'', we retrieved pre-trained models for 11 different tasks e.g.~such as \textit{classification}, \textit{segmentation} and \textit{image generation}. We also recorded various meta-data such as depth and frequency of included operations for each model, and manually categorized the variety of used training sets into 16 visually distinctive groups like \textit{natural scenes, medical ct, seismic,} or \textit{astronomy}. In total, the models were trained on 71 different datasets. The dominant subset is formed by \textit{image classification} models trained on \textit{ImageNet1k} \cite{imagenet} (355 models). 

\noindent All models were trained with full 32-bit precision\footnote{Although, initial experiments indicated that mixed/reduced precision training \cite{micikevicius2018mixed} does not affect distribution shifts beyond noise.} but may have been trained on variously scaled input data. 
Included in the dataset are low-resolution variants of AlexNet \cite{alexnet}, DenseNet-121/161/169 \cite{densenet}, ResNet-9/14/18/34/50/101/152 \cite{resnet}, VGG-11/13/16/19 \cite{vgg}, MobileNet v2 \cite{sandler2019mobilenetv2}, Inception v3 \cite{inceptionv3} and GoogLeNet \cite{inception}
image classification models that we have purposely trained on simple datasets such as CIFAR-10/100 \cite{cifar10}, MNIST \cite{mnist}, Kuzushiji-MNIST (KMNIST) \cite{kmnist} and Fashion-MNIST \cite{fashionmnist} in order to study the effect of overparameterization on learned filters.

\noindent All collected models were converted into the ONNX format \cite{bai2019} which allows a streamlined filter extraction without framework dependencies. Hereby, only the widely used filters from regular convolution layers with a kernel size of $3\times 3$ were taken into account. Transposed (sometimes also called de-convolution or up-convolution) convolution layers were not included. In total, \textbf{1,464,797,156 filters} from \textbf{21,436 layers} have been obtained for our dataset. 
\subsection{Analyzing filter structures}\label{subsec:structure}

\noindent We apply a full-rank principal component analysis (PCA) transformation implemented via a singular-value decomposition (SVD) to understand the underlying structure of the filters \cite{Jolliffe1986}.

\noindent First, we stack the relevant set of $n$ flattened filters into a $n\times 9$ matrix $X$. Thereupon, we center the matrix and perform a SVD into a ${n\times 9}$ rotation matrix $U$, a $9\times 9$ diagonal scaling matrix $\Sigma$, and a $9\times 9$ rotation matrix $V^{T}$. The diagonal entries $\sigma_{i}, i=0,\dotsc ,n-1$ of $\Sigma$ form the singular values in decreasing order of their magnitude. Row vectors $v_{i}, i=0,\dotsc ,n-1$ in $V^{T}$ then form the principal components. Every row vector $c_{ij}, j=0,\dotsc ,n-1$ in $U$ is the coefficient vector for $f_{i}$.
\begin{equation}
    \begin{split}
        X^{*} &= X - \bar{X} = U\Sigma V^{T} \\
    \end{split}
\end{equation}
Where $\bar{X}$ denotes the vector of column-wise mean values of any matrix $X$.
Then we obtain a vector $\hat{a}$ of the explained variance ratio of each principal component. $\displaystyle \|\cdot\|_{1}$ denotes the $\ell_{1}$-norm.
\begin{equation}
    \begin{split}
        \vec{a} &= {(\Sigma\cdot I)^{2}}/{(n - 1)} \\
        \hat{a} &= {\vec{a}}/{\|\vec{a}\|_{1}} 
    \end{split}
\end{equation}
Finally, each filter $f'$ is described by a linear, shifted sum of principal components $v_{i}$ weighted by the coefficients $c_{i}$.
\begin{equation}
    \begin{split}
        \displaystyle f'&=\sum_{i}c_{i}v_{i} + \bar{X}_{i}
    \end{split}
\end{equation}

\subsection{Measuring distribution shifts}


\noindent All probability distributions are represented by histograms. The histogram range is defined by the minimum and maximum value of all coefficients. Each histogram is divided into 70 uniform bins. The divergence between two distributions is measured by the symmetric, non-negative variant of Kullback-Leibler ($KL_{\text{sym}}$) \cite{kullback1951}. 
\begin{equation}
    \begin{split}
        \displaystyle 
        KL(P\parallel Q) &= \sum_{x\in {\mathcal {X}}}P(x)\log {\frac {P(x)}{Q(x)}}\\
        KL_\text{sym}(P\parallel Q) &= KL(P\parallel Q) + KL(Q\parallel P)
        \end{split}
\end{equation}
We define the drift $D$ between two filter sets by the sum of the divergence of the coefficient distributions $P_{i}, Q_{i}$ along every principal component index $i$. The sum is weighted by the ratio of variance $\hat{a}_{i}$ explained by the $i$-th principal component. 
\begin{equation}
    \begin{split}
        \displaystyle D(P\parallel Q)&=\sum_{i}\hat{a}_{i}\cdot KL_\text{sym}\left(P_{i}\parallel Q_{i}\right)
        \end{split}
\end{equation}
To avoid undefined expressions, all probability distributions $F$ are set to hold $\forall x\in {\mathcal {X}}: F(x)\geq\epsilon$.
\begin{figure*}
  \centering
  \begin{subfigure}{0.49\linewidth}
    \includegraphics[width=\linewidth]{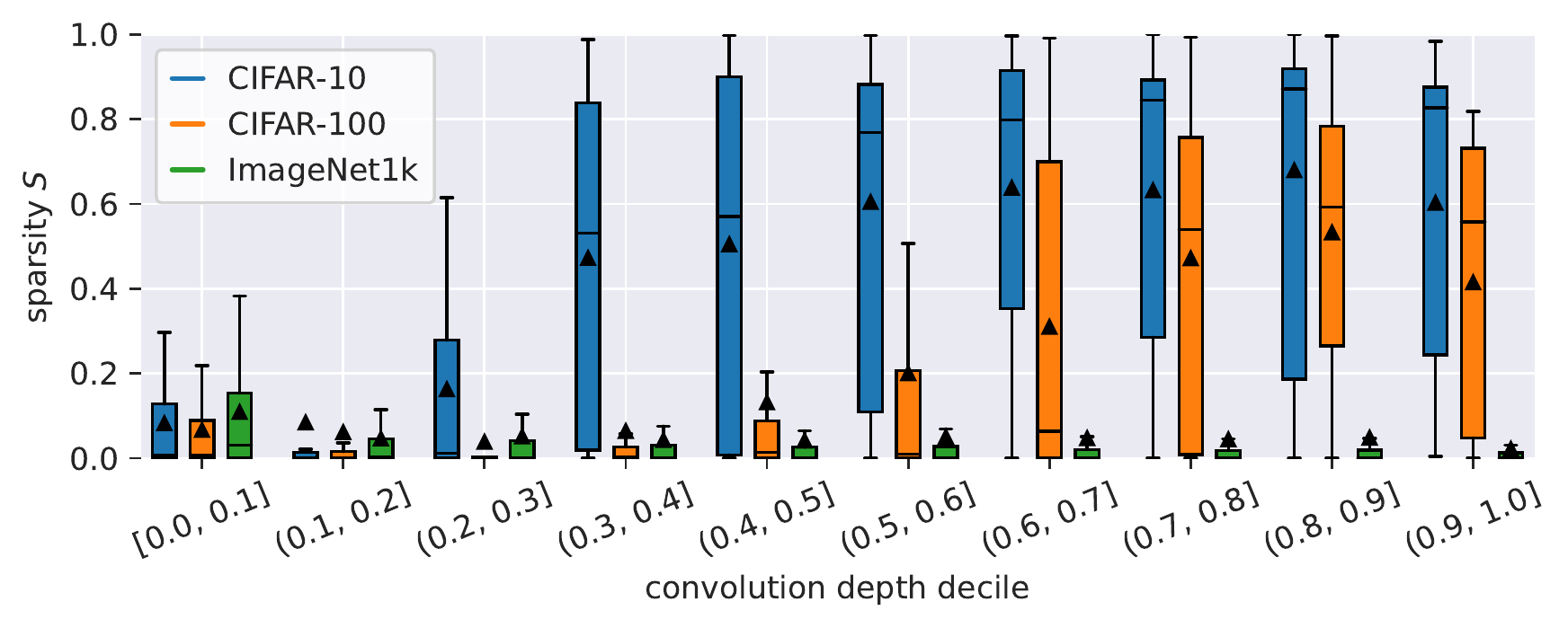}
    \caption{sparsity vs. overparameterization}
    \label{fig:degeneration_overparam_sparsity}
  \end{subfigure}
  \hfill
  \begin{subfigure}{0.49\linewidth}
    \includegraphics[width=\linewidth]{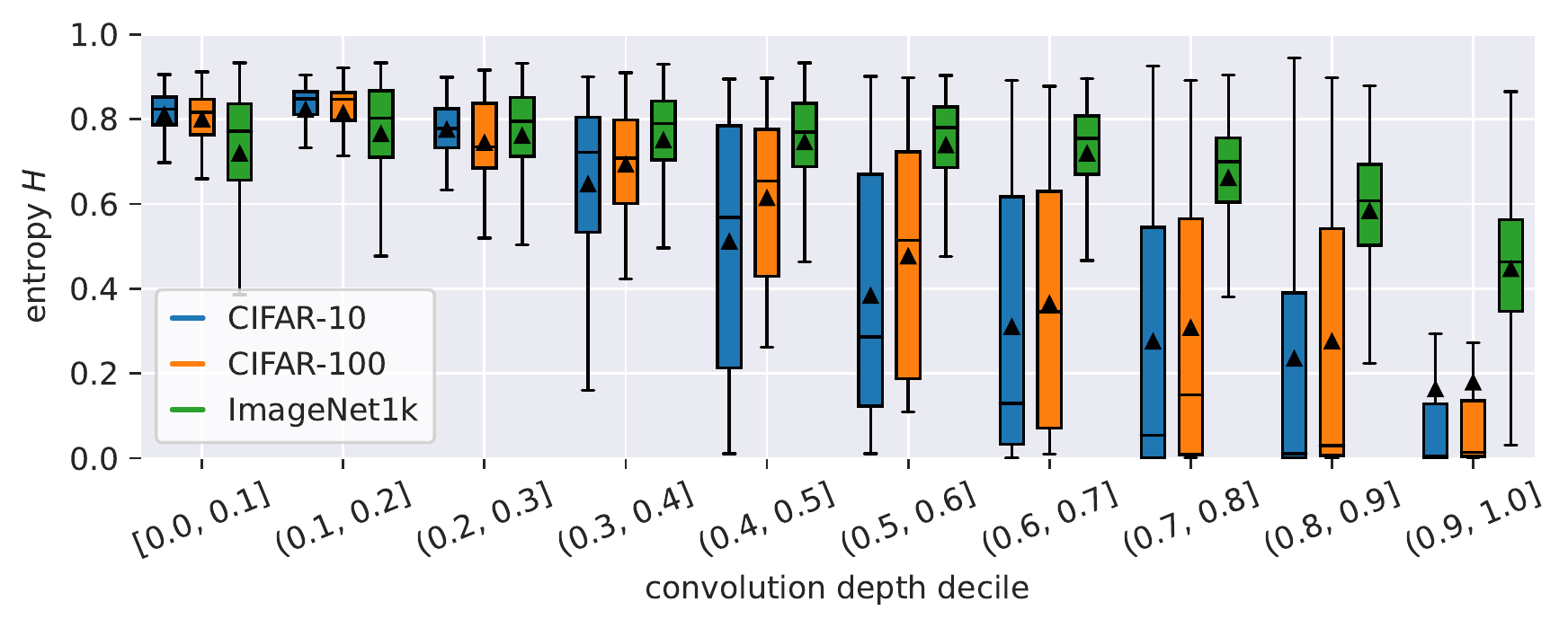}
    \caption{entropy vs. overparameterization}
    \label{fig:degeneration_overparam_entropy}
  \end{subfigure}%
  
  \begin{subfigure}{0.49\linewidth}
    \includegraphics[width=\linewidth]{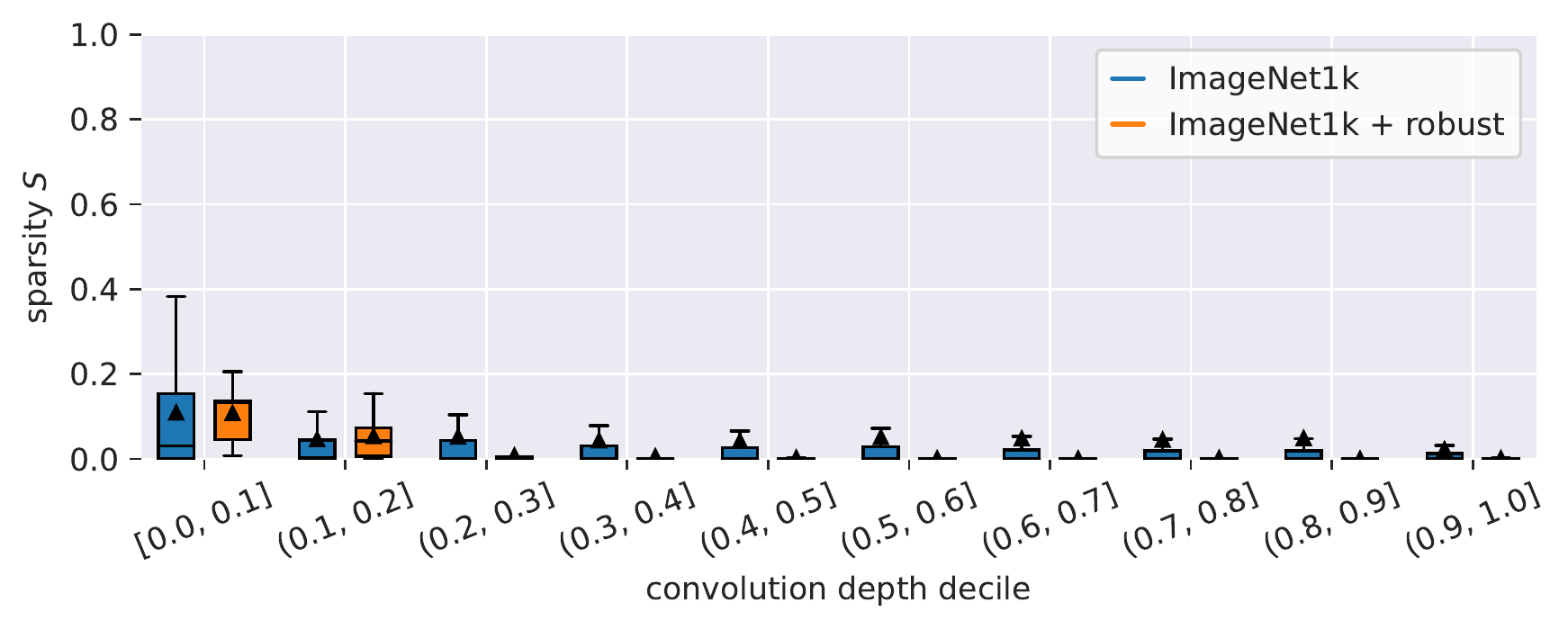}
    \caption{sparsity vs. robustness}
    \label{fig:degeneration_robust_sparsity}
  \end{subfigure}
  \hfill
  \begin{subfigure}{0.49\linewidth}
    \includegraphics[width=\linewidth]{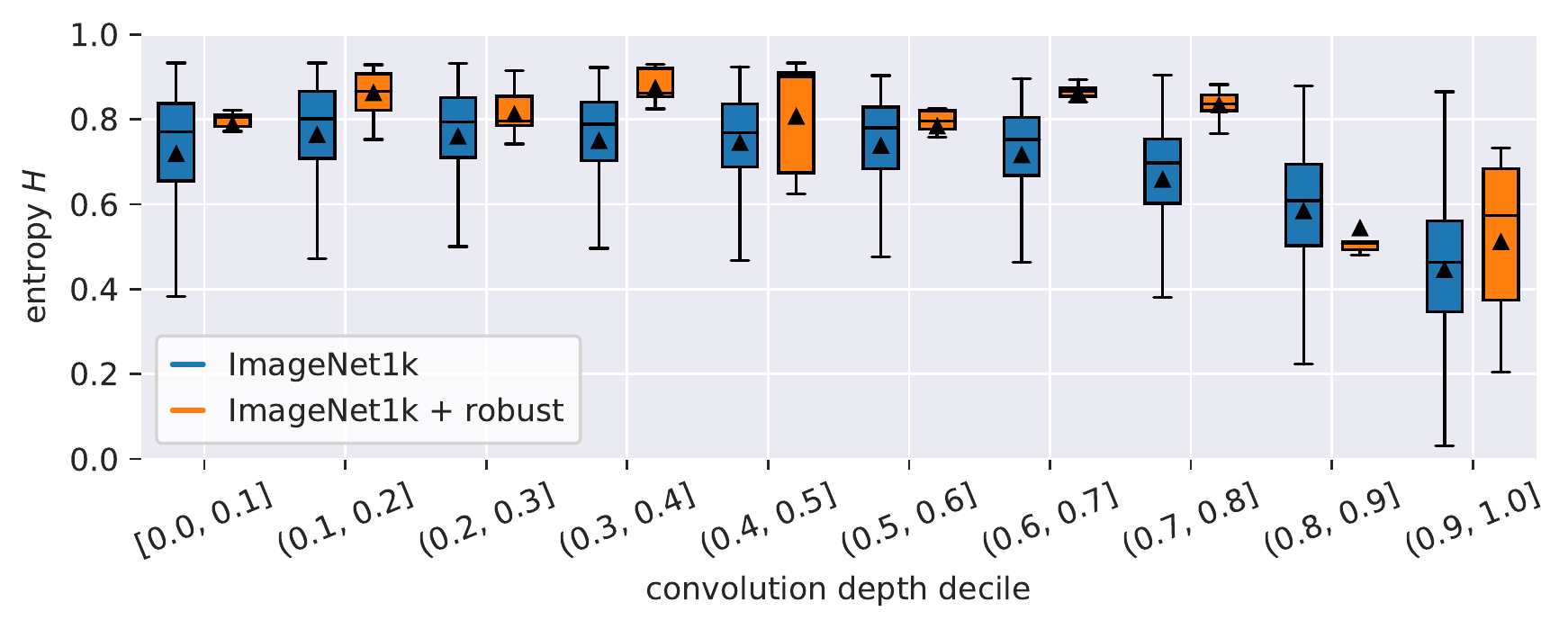}
    \caption{entropy vs. robustness}
    \label{fig:degeneration_robust_entropy}
  \end{subfigure}
  
  \caption{Comparison of layer entropy and sparsity of overparameterized, robust, and regular classification models. Outliers are hidden for clarity.}
  \label{fig:degeneration_overparam}
\end{figure*}
\subsection{Measuring layer degeneration}\label{subsec:layer_deg}
\noindent \textit{Lottery Ticket Hypothesis} \cite{frankle2018lottery} suggests that each architecture has a specific amount of convolution filters that saturate its ability to transform a given dataset into a well separable feature-space. Exceeding this number will result in a partitioning of the model into multiple inter-connected submodels. We hypothesize that these are seen in the form of degenerated filters in CNNs. In like manner, an insufficient amount of training samples or training epochs will also lead to degenerated filters. We characterize the following types of degeneration.
\begin{enumerate}
    \item \textit{High sparsity:} Filters are dominantly close to zero and therefore produce quasi-zero feature-maps \cite{8953939}. These feature-maps carry no vital information and can be discarded.
    \item \textit{Low diversity in structure:} Filters are structurally similar to each other and therefore redundant. They produce similar feature-maps in different scales and could be represented by a subset of present filters.
    \item \textit{Randomness:} Filter weights are conditionally independent of their neighbours. This indicates that no or not sufficient training was performed.
\end{enumerate}

\noindent Sparsity degeneration is detectable by the share of sparse filters $S$ in a given layer. We call a filter $f$ sparse if all entries are near-zero. Consequently, given the number of input channels $c_{\text{in}}$, number of output channels $c_{\text{out}}$, and a set of filters in layer $L$, we can measure the layer sparsity by:
\begin{equation}
    \begin{split}
        \displaystyle S(L)=\frac{|\{f | f \in L \land (\forall w \in f: -\epsilon_{0} \leq w \leq \epsilon_{0})\}|}{c_{\text{in}}c_{\text{out}}}
    \end{split}
\end{equation}
To detect the other types of degeneration we introduce a layer-wise metric based on the Shannon-Entropy of the explained variance ratio of each principal component obtained from a SVD of all filters in the examined layer (\cref{subsec:structure}). 
\begin{equation}
    {\displaystyle H=-\sum _{i}{\hat{a}_{i}\log_{10} \hat{a}_{i}}}
\end{equation}
If $\displaystyle H$ is close to zero this indicates one strong principal component from which most of the filters can be reconstructed and is therefore a low filter diversity degeneration. On the other hand, a large entropy indicates a (close to) uniform distribution of the singular values and, thus, a randomness of the filters. Sparse layers are a specific form of low diversity degeneration and, generally both are correlated, whereas, sparsity and randomness are mutually exclusive. It should be noted, that $|\Sigma\cdot I|=\min(c_{\text{in}}c_{\text{out}}, 9)$ and therefore the entropy only becomes expressive if $c_{\text{in}}c_{\text{out}} \gg 9$. 

%% file: sec/4_results.tex
\section{Results: Analysis of trained CNN filters }
\label{sec:results}
\subsection{Layer degeneration}

\noindent In this section we study different causes of degeneration and aim provide thresholds for evaluation.

\paragraph{Overparameterization}
The majority of the models that we have trained on our low resolution datasets are heavily overparameterized for these relatively simple problems. We base this argument on the fact that we have models with different depth for most architectures and already observe near perfect performance with the smallest variants. Therefore it is safe to assume that larger models are overparameterized especially given that the performance only increases marginally\textsuperscript{\ref{supp}}.

\noindent First we analyze layer sparsity and entropy for these models trained on \textit{CIFAR-10/100} in comparison to all \textit{ImageNet1k} classification models found in our dataset. For each dataset we have trained identical networks with identical hyper-parameters. Both, \textit{CIFAR-10} and \textit{CIFAR-100}, consist of 60,000 $32\times 32$px images, but \textit{CIFAR-100} includes 10x more labels and thus fewer samples per class forming a more challenging dataset.

\noindent\cref{fig:degeneration_overparam_sparsity} shows that the overparameterized models contain significantly more sparse filters on average, and that sparsity increases with depth. In particular, we see the most sparse filters for \textit{CIFAR-10}. However, \textit{ImageNet1k} classifiers also seem to have some kind of ``natural'' sparsity, even though we do not consider most of these models as overparameterized. Entropy, on the other hand, decreases with increasing layer depth for every \textit{classifier}, but more rapidly in overparameterized models (\cref{fig:degeneration_overparam_entropy}). Again, the \textit{CIFAR-10} models degrade faster and show more degeneration. 

\noindent The overparameterized models contain layers that have a entropy close to 0 towards deeper layers which indicates that these models are ``saturated'' and only produce differently scaled variants of the same filters. In line with the oversaturation, these models also have increasingly sparse filters, presumably as an effect of regularization.



\begin{figure*}
  \centering
  \includegraphics[width=\linewidth]{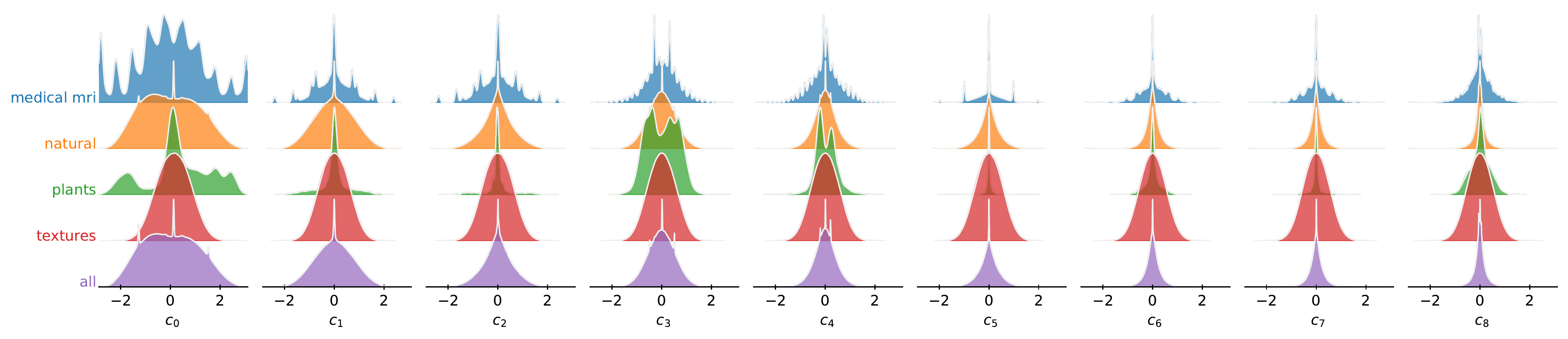}
  \caption{KDEs of the coefficient distributions along every principal component for selected\textsuperscript{\ref{supp}} visual categories.}
  \label{fig:ridge_selected}
\end{figure*}

\paragraph{Filter degeneration and model robustness}
Our dataset also contains robust models from the \textit{RobustBench leaderboard} \cite{croce2021robustbench}. When comparing robust models with non-robust models trained on \textit{ImageNet1k}, it becomes clear that robust models form almost no sparse filters after in deeper convolution layers (\cref{fig:degeneration_robust_sparsity}), while regular models show some sparsity there. The entropy of robust models is also higher throughout depth (\cref{fig:degeneration_robust_entropy}), indicating that robust models learn more diverse filters.

\paragraph{Thresholds}
To obtain a threshold for randomness given a number of filters $n$ per layer we perform multiple experiments in which we initialize convolution filters of different sizes from a standard normal distribution and fit a sigmoid $T_H$ to the minimum results obtained for entropy.
\begin{equation}
        \displaystyle T_H(n)=\frac{L}{1+e^{-k(\log_{2}(n)-x_{0})}}+b
\end{equation}
We obtain the following values $L=1.26$, $x_0=2.30$, $k=0.89$, $b=-0.31$ and call any layer $L$ with $H > T_H(n)$ random. 
\noindent On the opposite, defining a threshold for low diversity degeneration seems less intuitive and one can only rely on statistics: The average entropy $H$ is 0.69 over all layers and continuously decreases from an average of 0.75 to 0.5 with depth. Additionally, the minimum of the 1.5 IQR also steadily decreases with depth. 

\noindent The same applies to sparsity: the average sparsity $S$ over all layers is 0.12 and only 56.5\% of the layers in our dataset hold $S < 0.01$ and 9.9\% even show $S > 0.5$. In terms of convolution depth, the average sparsity varies between 9.9\% and 14\% with the largest sparsity found in the last 20\% of the model depth. The largest outliers of the 1.5 interquartile range (IQR) are, however, found in the first decile. 
In both cases we find it difficult to provide a meaningful general threshold and suggest to determine this value on a case-by-case basis\textsuperscript{\ref{supp}}.%
\subsection{Filter structure}
\begin{figure*}
    \begin{minipage}{\columnwidth}\vspace{0pt}%
        \centering
        \begin{tabular}{m{0.95\linewidth}}
            \textcolor{colortab0}{$\bullet$\hspace{5pt}} \adjustbox{valign=c}{\includegraphics[width=0.9\linewidth]{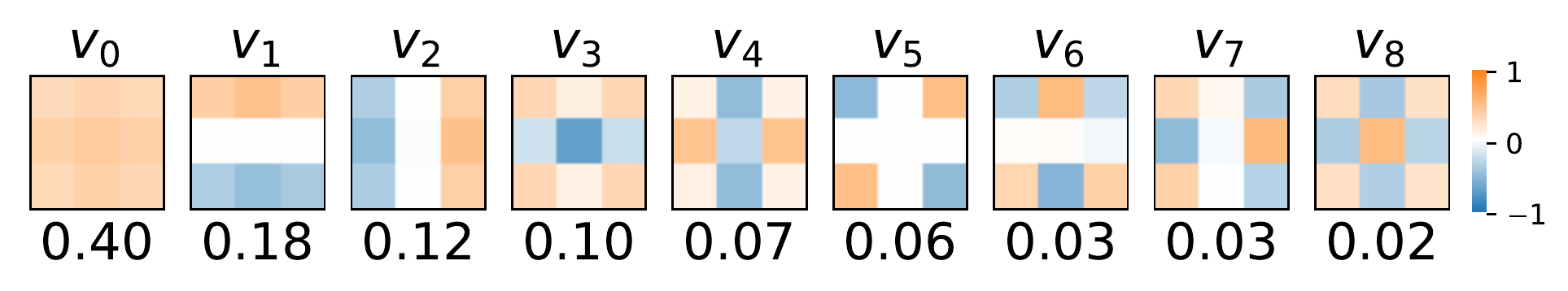}} \\ 
            \textcolor{colortab1}{$\bullet$\hspace{5pt}} \adjustbox{valign=c}{\includegraphics[width=0.9\linewidth]{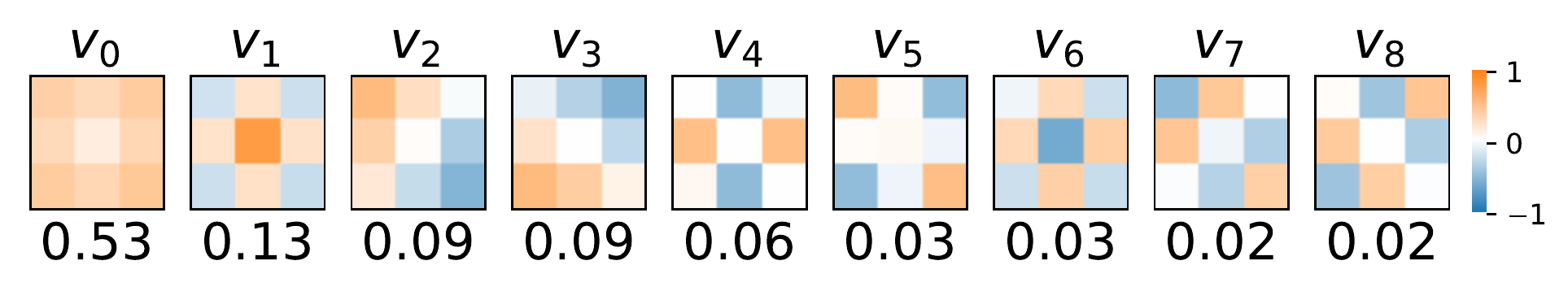}}\\
            \textcolor{colortab2}{$\bullet$\hspace{5pt}} \adjustbox{valign=c}{\includegraphics[width=0.9\linewidth]{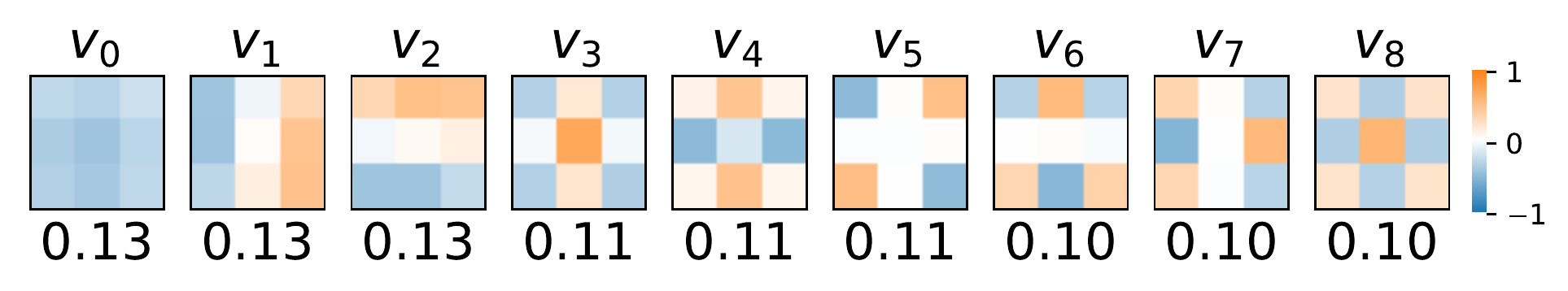}}\\
            \textcolor{colortab3}{$\bullet$\hspace{5pt}} \adjustbox{valign=c}{\includegraphics[width=0.9\linewidth]{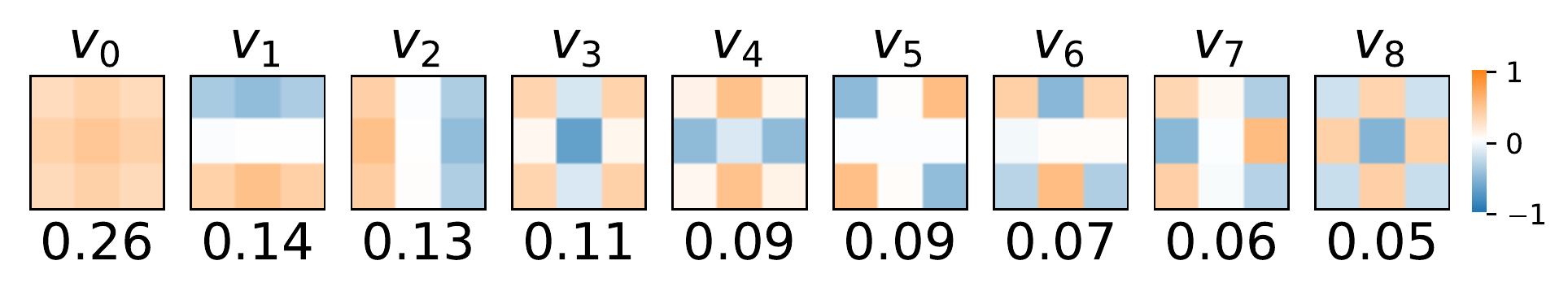}}
        \end{tabular}
    \end{minipage}\hspace{20pt}%
    \begin{minipage}{\columnwidth}%
        \centering
        \includegraphics[width=0.95\linewidth]{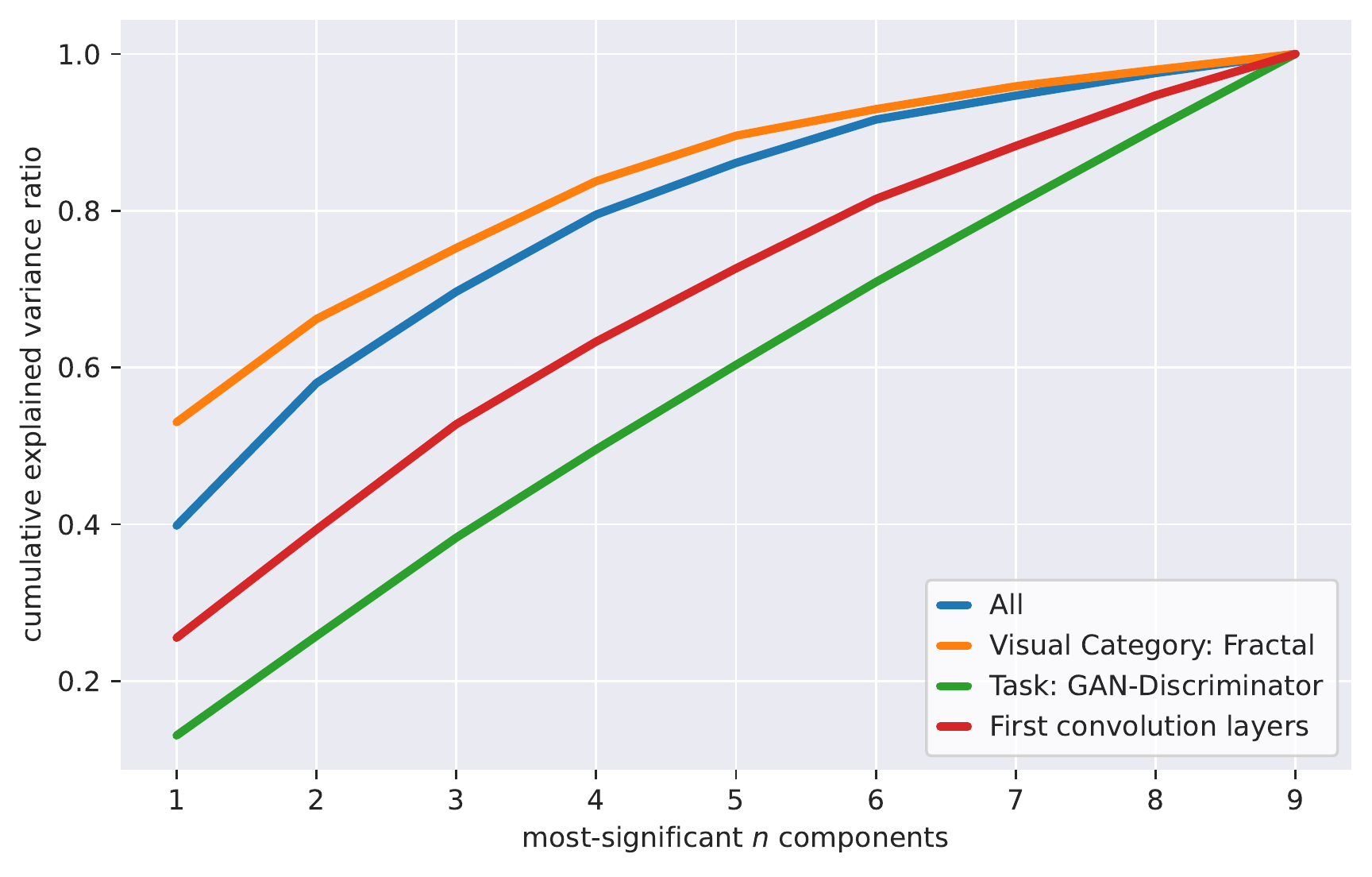}
    \end{minipage}
    \caption{Selected\textsuperscript{\ref{supp}} depiction of the filter basis and (cumulative) explained variance ratio per component for filters from \textcolor{colortab0}{$\bullet$}~full dataset, \textcolor{colortab1}{$\bullet$}~models trained on images of \textit{fractals}, \textcolor{colortab2}{$\bullet$}~GAN discriminators, \textcolor{colortab3}{$\bullet$}~first convolution layers.}
    \label{fig:filter_basis}
\end{figure*}

\begin{figure}
  \centering
  \includegraphics[width=\linewidth]{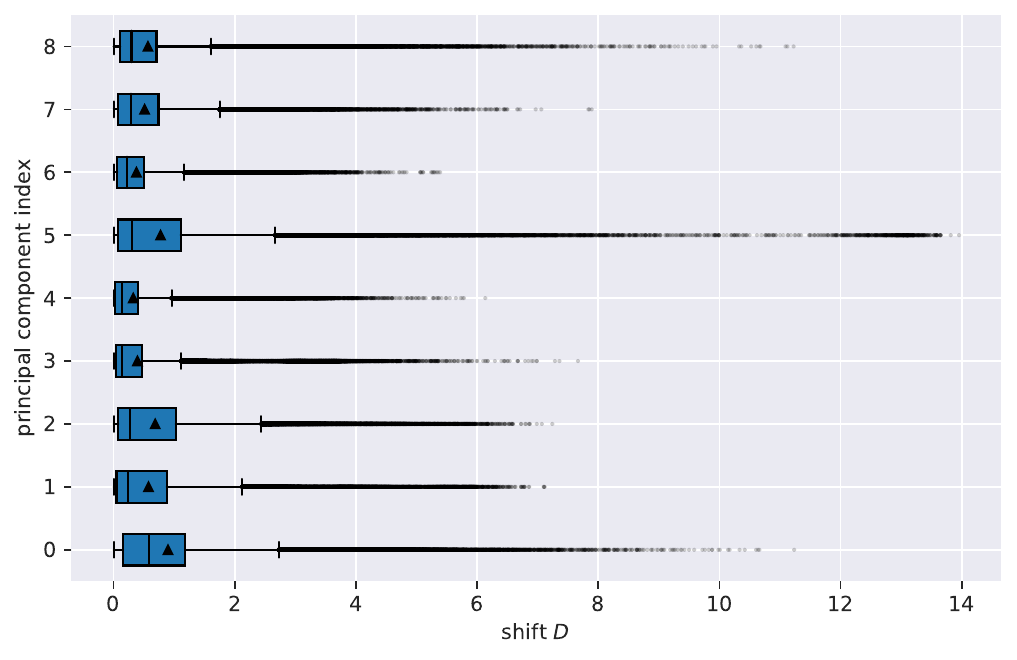}
  \caption{Distribution of the shift $D$ along principal components computed on all possible pairings of models.}
  \label{fig:kl_dimensions}
\end{figure}

\noindent In the next series of experiments, we analyze only the structure of $3\times 3$ filters, neglecting their actual numerical weight in the trained models. Therefore, we normalize each filter $f$ individually by the absolute maximum weight into $f'$.
\begin{equation}
    \begin{split}
        \displaystyle d_i &= \max_{i,j} \left|f_{ij}\right|\\
        f_{ij}' &= \left. 
        \begin{cases}
            f_{ij} / d_i, & \text{if } d_i \neq 0 \\
            f_{ij}, & \text{else} 
      \end{cases}
      \right.\\
    \end{split}
\end{equation}
Then we perform a PCA transformation on the scaled filters. \cref{fig:filter_basis} shows some qualitative examples of obtained principal components, split by several meta-data dimensions. The images of the formed basis are often similar for all groups except for few outliers (such as \textit{GAN-discriminators}). The explained variance however fluctuates significantly and sometimes changes the order of components. Consistently, we observe substantially higher variance on the first principal components. The explained variance does not necessarily correlate with the shift observed between models. Here, the biggest mean drift is also located in the first principal component ($\hat{D}=0.90$), but is then followed by the sixth, third, second component ($\hat{D}=0.78, 0.69, 0.58$). The coefficients of the sixth component also contain the strongest outliers (\cref{fig:kl_dimensions}).
\begin{figure}
    \centering
    
    \begin{subfigure}{0.24\columnwidth}
        \includegraphics[width=\linewidth]{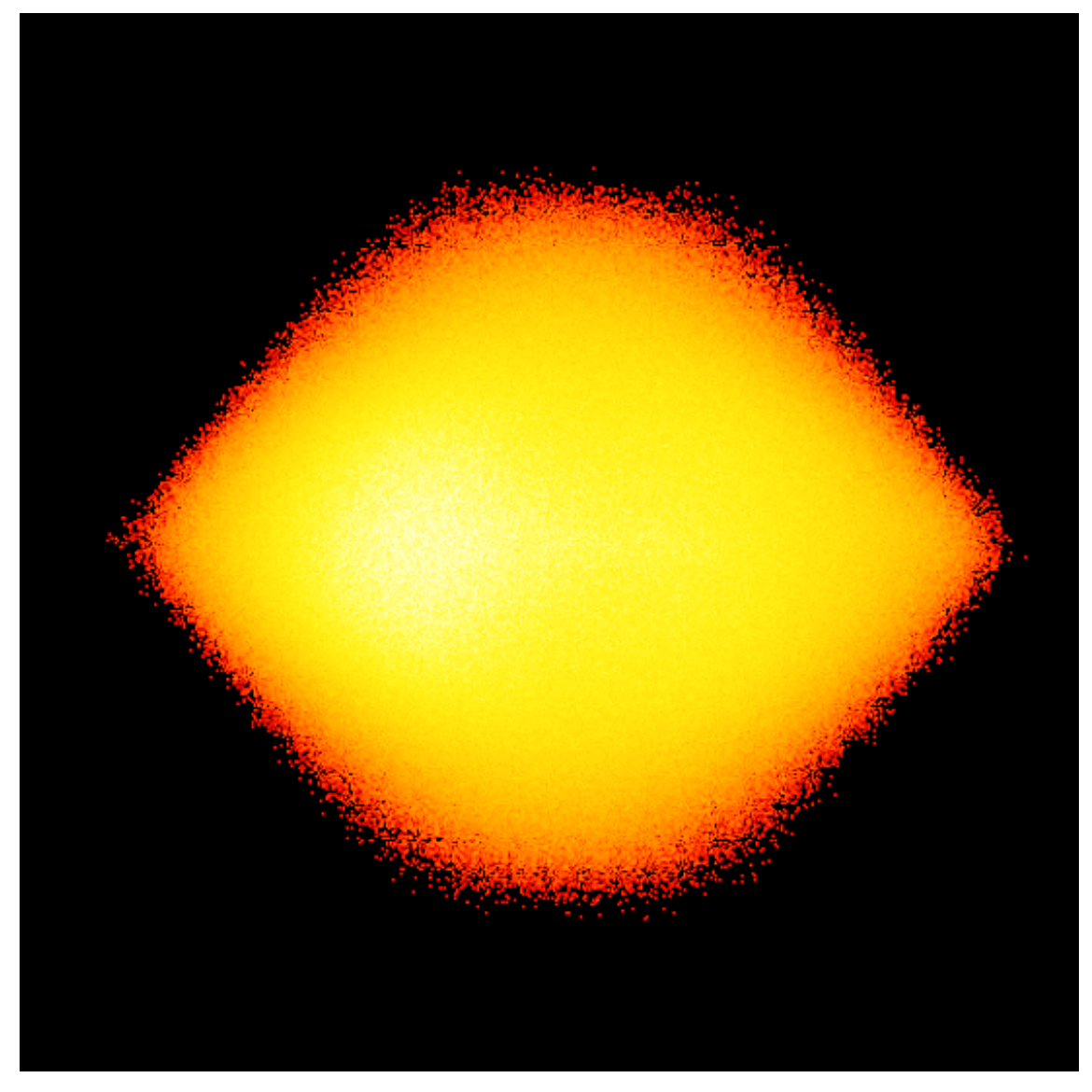}
        \caption{sun}
    \end{subfigure}
    \begin{subfigure}{0.24\columnwidth}
        \includegraphics[width=\linewidth]{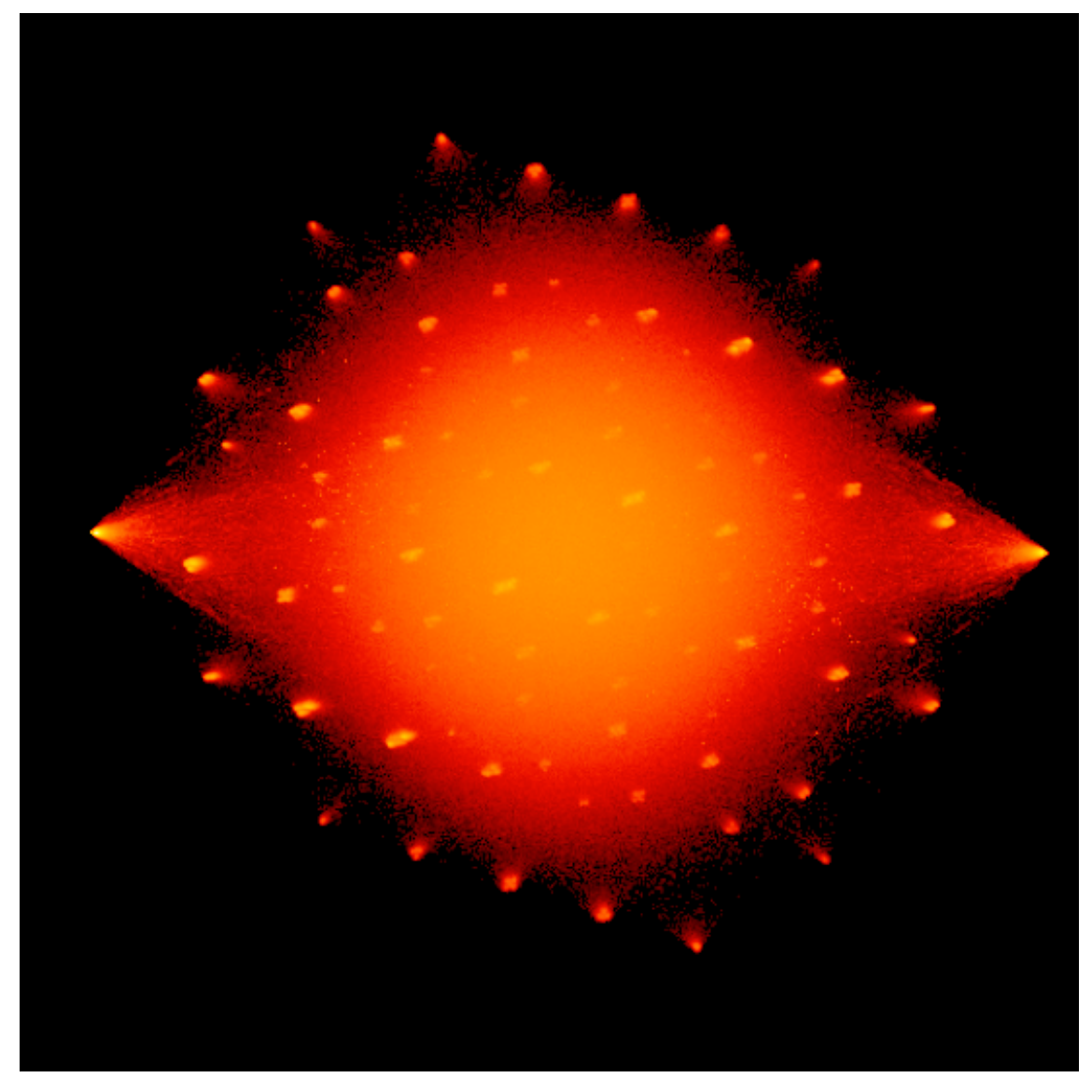}
        \caption{spikes}
    \end{subfigure}
    \begin{subfigure}{0.24\columnwidth}
        \includegraphics[width=\linewidth]{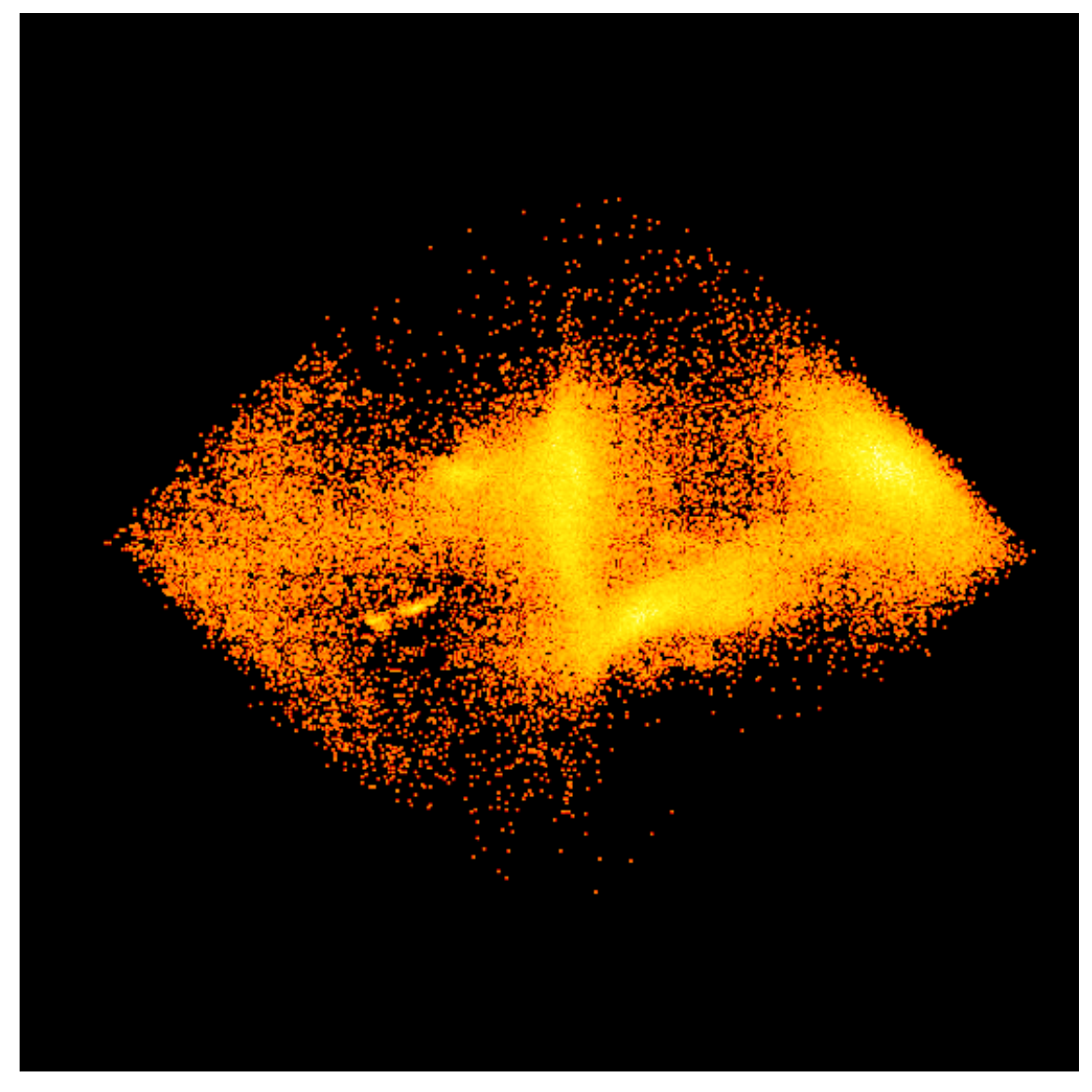}
        \caption{symbols}
    \end{subfigure}
    \begin{subfigure}{0.24\columnwidth}
        \includegraphics[width=\linewidth]{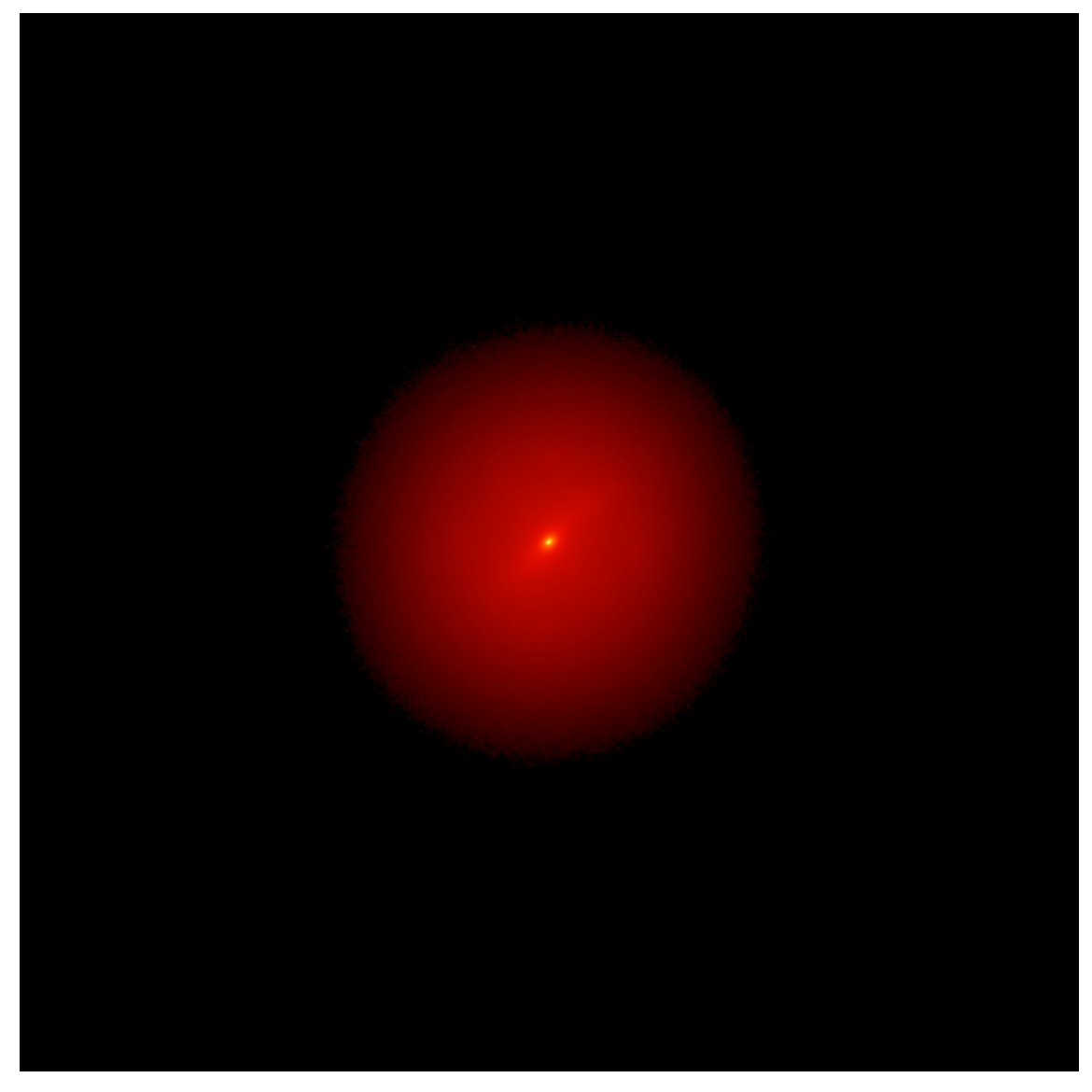}
        \caption{point}
    \end{subfigure}

    \caption{Bi-variate plots between component distributions showing the four phenotypes.}
    \label{fig:scatter_pheno}
\end{figure}
\noindent We visualize the distributions of PCA coefficients along every component for each group by plots of kernel density estimates (KDEs), e.g. \cref{fig:ridge_selected} depicts the distributions of filters grouped by some selected visual categories in comparison to the distribution of coefficients for the full dataset. Filters extracted from models with degenerated layers (as seen in \textit{medical mri}) result in spiky/multi-modal KDEs. The distributions can alternatively be visualized by bi-variate scatter plots that may reveal more details than KDEs. For example, they let us categorize the distributions into phenotypes depending on their distribution characteristic in the PCA space (\cref{fig:scatter_pheno}): \textit{sun:} distributions where both dimensions are gaussian-like. These are to be expected coefficient distributions without significant sparsity/low diversity degeneration. Yet, this phenotype may also include non-converged filters; \textit{spikes:} distributions suffering from a low variance degeneration resulting in local hotspots; \textit{symbols:} at least one distribution is multi-modal, non-centered, highly sparse or otherwise non-normal (low variance degeneration); \textit{point:} coefficients are primarily located in the center (sparsity degeneration).
%
%

\paragraph{Reproducibility of filters}
We train low-resolution networks on \textit{CIFAR-10} multiple times with identical hyper-parameters except for random seeds and save a checkpoints of each model at the best validation epoch. 
Most models are converging to highly similar coefficient distributions when retrained with different weight initialization (e.g. ResNet-9 with $D < 5.3\cdot 10^{-4}$). However, some architectures such as \textit{MobileNetv2} show higher shifts ($D < 2.6\cdot 10^{-2}$). We assume that this is due to the structure of the loss surface, e.g. the residual skip connections found in \textit{ResNets} smooth the surface, whereas other networks way contain more local minima due to noisy surfaces \cite{wang2020skip}.
\begin{figure}
  \centering
  \includegraphics[width=\columnwidth]{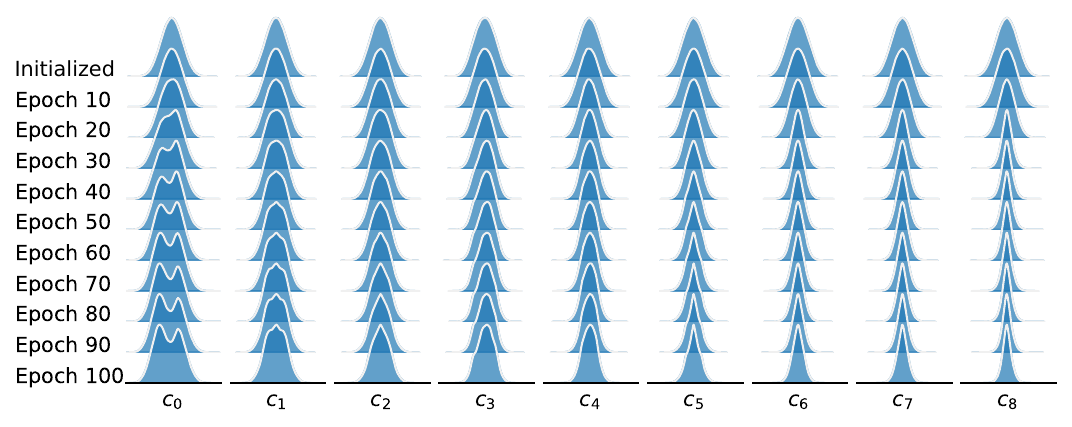}
  \caption{Coefficient distribution of a \textit{ResNet-9} trained on \textit{CIFAR-10} every 10 epochs.}
  \label{fig:ridge_resnet9_in_training}
\end{figure}
%

\paragraph{Formation of filter structures during training}
Although our dataset only includes trained convolutional filters we tried to understand how the coefficient distribution shifts during training. Therefore we recorded checkpoints of a \textit{ResNet-9} trained on \textit{CIFAR-10} every 10 training epochs beginning right after the weight initialization. \cref{fig:ridge_resnet9_in_training} shows that the coefficient distributions along all principal components are gaussian-like distributed in the beginning and eventually shift during training. For this specific model, distributions along major principal components retain the standard deviation during training, while less-significant component distributions decrease.
The initialization observation helped us removing models from our collection where we failed to load the trained parameters for any reason and is foundation for our provided randomness metric.
\begin{figure*}
  \centering
  \includegraphics[width=\linewidth]{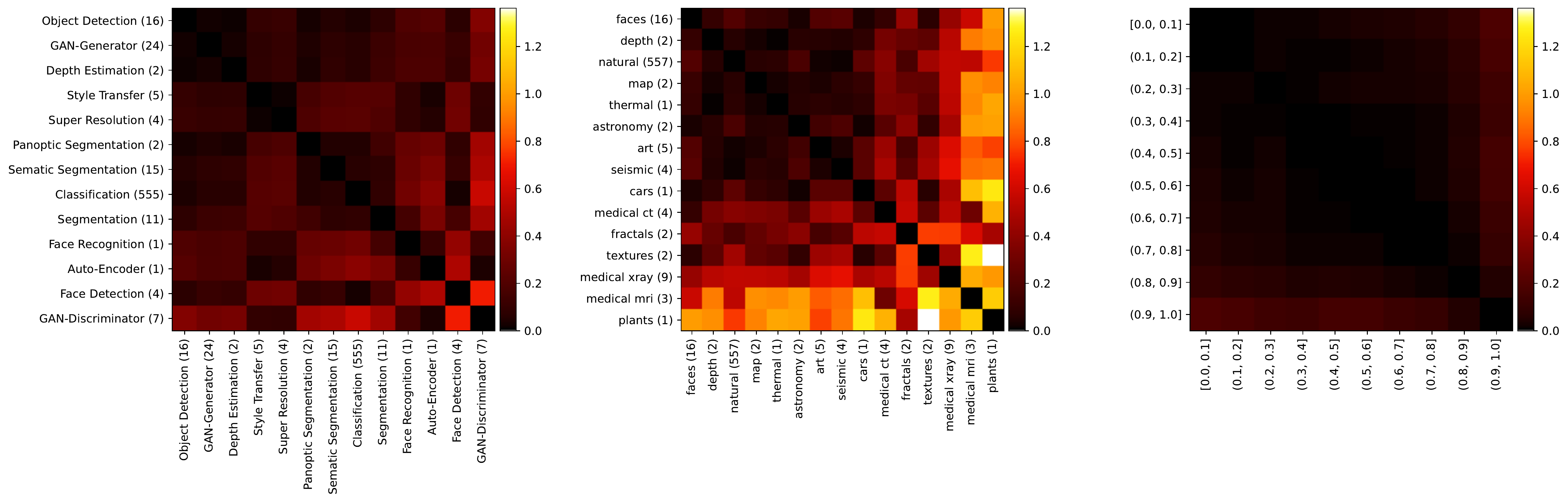}
  \begin{tabularx}{\linewidth}{>{\centering\arraybackslash}X>{\centering\arraybackslash}X>{\centering\arraybackslash}X}
        (a) tasks & (b) visual categories & (c) convolution depth decile\\
  \end{tabularx}
  \caption{Heatmaps over the shift $D$ for different filters groupings. The number in brackets denotes the number of models in this group. Low values/dark colors denote low shifts.}
  \label{fig:kl_combined}
\end{figure*}
\subsection{Distribution shifts between trained models}
\noindent In this subsection we are investigating transfer distance in different meta-dimensions of pre-trained models. We compute the shift $D$ and visualize this is the form of heatmaps (\cref{fig:kl_combined}) that show shifts between all pairings.
%

\paragraph{Shifts between tasks}
Unsurprisingly, \textit{classification}, \textit{segmentation}, \textit{object detection}, and \textit{GAN-generator} distributions are quite similar, since the non-\textit{classification} models typically include a \textit{classification} backbone. The smallest mean shift to other tasks is observed in \textit{object detection}, \textit{GAN-generators}, and \textit{depth estimation} models. 
The least transferable distributions are \textit{GAN-discriminators}. Their distributions do barely differ along principal components and can be approximated by a gaussian distribution. By our randomness metric this indicates a filter distribution that is close to random initialization, implying a ``confused'' \textit{discriminator} that cannot distinguish between real and fake samples towards the end of (successful) training.
It may be surprising to see a slightly larger average shift for \textit{classification}. This is presumably due to many degenerated layers in our collected models, which are also visible in the form of spikes when studying the KDEs. An evaluation\textsuperscript{\ref{supp}} of distributions including only non-degenerated \textit{classifiers} actually shows a lower average shift due to the aforementioned similarity to other tasks.
%

\paragraph{Shifts between visual categories and training sets}
We find that the distribution shift is well balanced across most visual categories and training sets. Notable outliers include all \textit{medical} types. They have visible spikes in the KDEs, once again indicating degenerated layers. Indeed, the average sparsity in these models is extreme in the last 80\% of the model depth. 
Another interesting, albeit less significant outlier is the \textit{fractal} category. It consists of models trained on \textit{Fractal-DB}, which was proposed as a synthetic pre-training alternative to \textit{ImageNet1k} \cite{KataokaACCV2020}. The standard deviations of coefficient distributions tend to shrink towards the least significant principal components but this trend is not visible for this category indicating that sorting the basis by variance would yield a different order for this task and perhaps the basis itself is not well suited. Also notable is a remarkably high standard deviation on the distribution of the first principal component. Interestingly, we also observe sub-average degeneration for this category.
Shifts in other categories can usually be explained by a biased representation. For example we only have one model for \textit{plants}, our \textit{handwriting} models consist exclusively of overparameterized networks that suffer from layer degeneration, and \textit{textures} consists of only one \textit{GAN-discriminator} which will naturally shows a high randomness.
\begin{figure}
  \centering
  \includegraphics[width=\columnwidth]{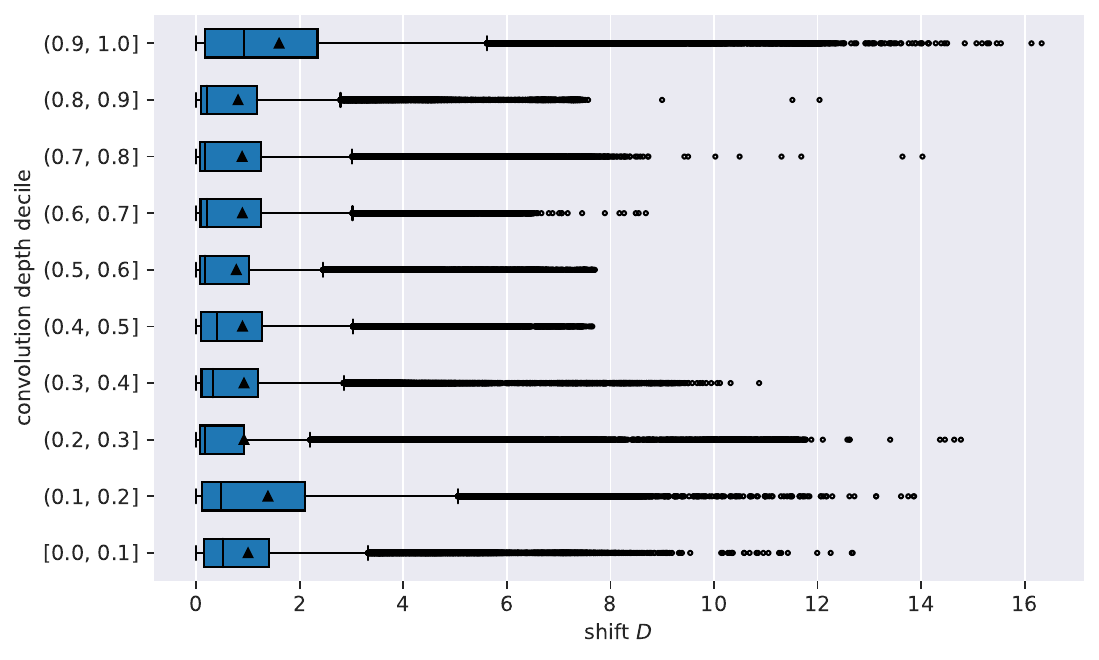}
  \caption{Boxplots showing the distribution of pair-wise model-to-model shift $D$ of \textit{classification} models per convolution depth decile (top to bottom in decreasing order). Our intentionally overparameterized models were left out of this analysis.}
  \label{fig:boxplot_kl_conv_depth_norm_classifier_comparison_no_fliers}
\end{figure}%
%

\paragraph{Shifts by filter/layer depth}
The shift between layers of various depth deciles increases with the difference in depth, with distributions in the last decile of depth forming the most distinct interval, and outdistancing the second-to-last and first decile that follow next.
An interesting aspect is also the model-to-model shift across deciles. This shift exemplifies the uniqueness of formed filters. Our observations overhaul the general recommendation for fine-tuning to freeze early layers in \textit{classification} models, as the largest shifts are not only seen in deep layers but also in early vision (\cref{fig:boxplot_kl_conv_depth_norm_classifier_comparison_no_fliers}). \textit{Segmentation}\textsuperscript{\ref{supp}} models show the most drift in deeper layers.
Contrary, \textit{object/face detection} models only show drift in the early vision (\textit{object detection} in the first, \textit{face detection} in the first four depth deciles), but marginal drift in later convolution stages.
%

\begin{figure}
  \centering
  \includegraphics[width=\columnwidth]{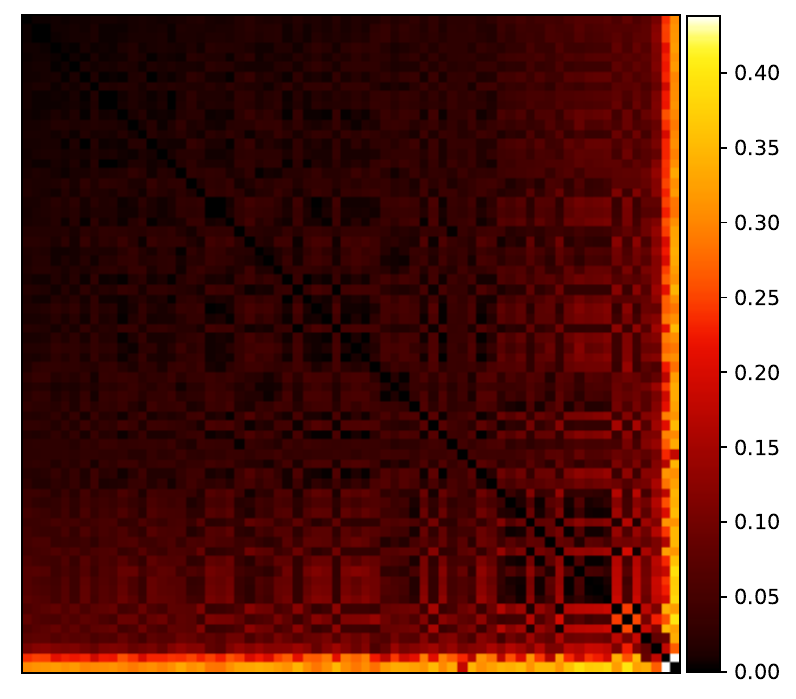}
  \caption{Heatmap over the shift $D$ between different pairings of \textit{ResNet-classifiers}. Each row/column depicts one model. Intentionally overparameterized models were not included.}
  \label{fig:kl_resnet}
\end{figure}

\paragraph{Shifts within model families} The shift between models of the same family trained for the same task is negligible (\cref{fig:kl_resnet}), indicating that every large enough dataset is good enough and the common practice of pre-training models with \textit{ImageNet1k} even for visually distant application domains is indeed a valid approach. \textit{ResNet}-family outliers only consist out of models that show a high amount of sparsity. Additionally, this observation may be exploited by training small teacher networks and apply knowledge distillation\cite{hinton2015distilling} to initialize deeper models of the same family.\\

%% file: sec/limitations.tex
\section{Limitations}\label{sec:limitations}
\noindent Our data is biased against \textit{classification} models and/or \textit{natural} datasets such as \textit{ImageNet1k}. Further, some splits will over-represent specific dimensions e.g.~tasks may include exclusive visual categories and vice versa. Also, as previously shown, many of the collected models show a large amount of degenerated layers that impact the distributions. This also biases measurements of the distribution shifts.
We performed an ablation study by removing filters extracted from degenerated layers, but were unable to find a clear correlation between degeneration and distribution shifts\textsuperscript{\ref{supp}}, presumably due to a lack of justified thresholds.

%% file: sec/5_conclusions.tex
\section{Conclusions}
\label{sec:conclusions}
\noindent Our first results support our initial hypothesis that the distributions of trained convolutional filters are a suitable and easy-to-access proxy for the investigation of image distributions in the context of transferring pre-trained models and robustness. While the presented results are still in the early stages of a thorough study, we report several interesting findings that could be explored to obtain better model generalizations and assist in finding suitable pre-trained models for fine-tuning.   
One finding is the presence of large amounts of degenerated (or untrained) filters in large, well-performing networks - resulting in the phenotypes \textit{points}, \textit{spikes}, and \textit{symbols}. We assume that their existence is a symptom in line with the \textit{Lottery Ticket Hypothesis} \cite{frankle2018lottery}. We conclude that ideal models should have relatively high entropy (but $H < T_{H}$) throughout all layers and almost no sparse filters. Models that show an increasing or generally high sparsity or a massive surge in entropy with depth are most likely overparameterized and could be pruned, which would benefit inference and training speed. Whereas, initialized but not trained models will have a constantly high entropy $H \geq T_{H}$ throughout all layers and virtually no sparsity.\\
Another striking finding is the observation of very low shifts in filter structure between different meta-groups: I) shifts inside a family of architectures are very low; II) shifts are mostly independent of the target image distribution and task; III) also we observe rather small shifts between convolution layers of different depths with the highest shifts in the first and last layers.
Overall, the analysis of over 1.4 billion learned convolutional filters in the provided dataset gives a strong indication that the common practice of pre-training CNNs is indeed a sufficient approach if the chosen model is not heavily overparameterized.
Our first results indicate that the presented dataset is a rich source for further research in transfer learning, robustness and pruning.   

%% file: sec/X_supplementary.tex
\appendix

\setcounter{page}{1}

\twocolumn[
\centering
\Large
\textbf{CNN Filter DB: An Empirical Investigation of Trained Convolutional Filters} \\
\vspace{0.5em}Supplementary Material \\
\vspace{1.0em}
] 
\appendix

\begin{figure*}
  \centering
  
  \begin{subfigure}{0.49\linewidth}
    \includegraphics[width=\linewidth]{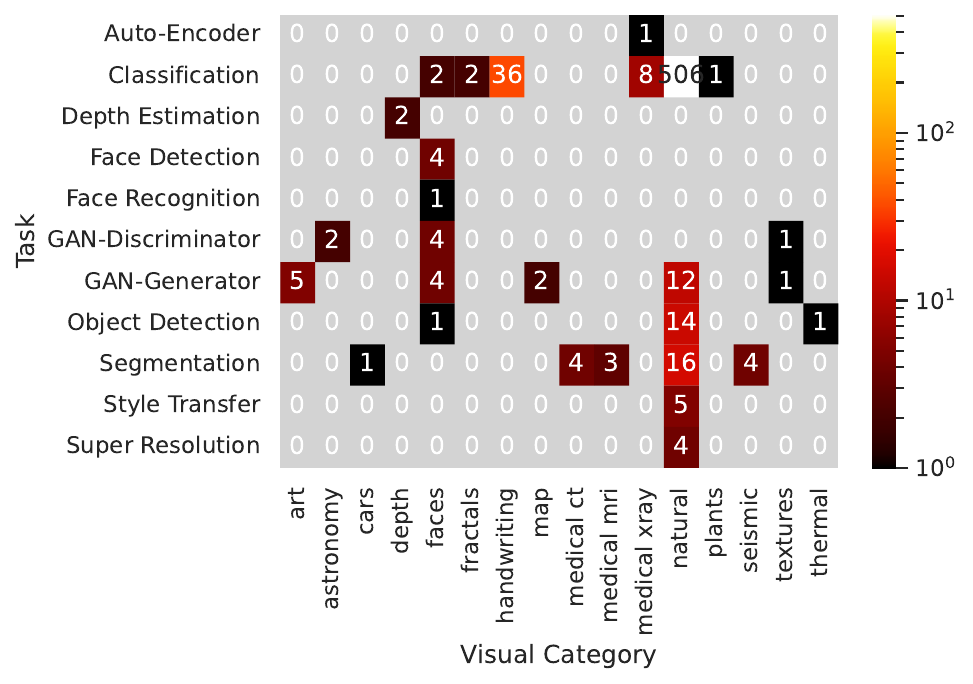}
    \caption{model frequency}
  \end{subfigure}
  \begin{subfigure}{0.49\linewidth}
    \includegraphics[width=\linewidth]{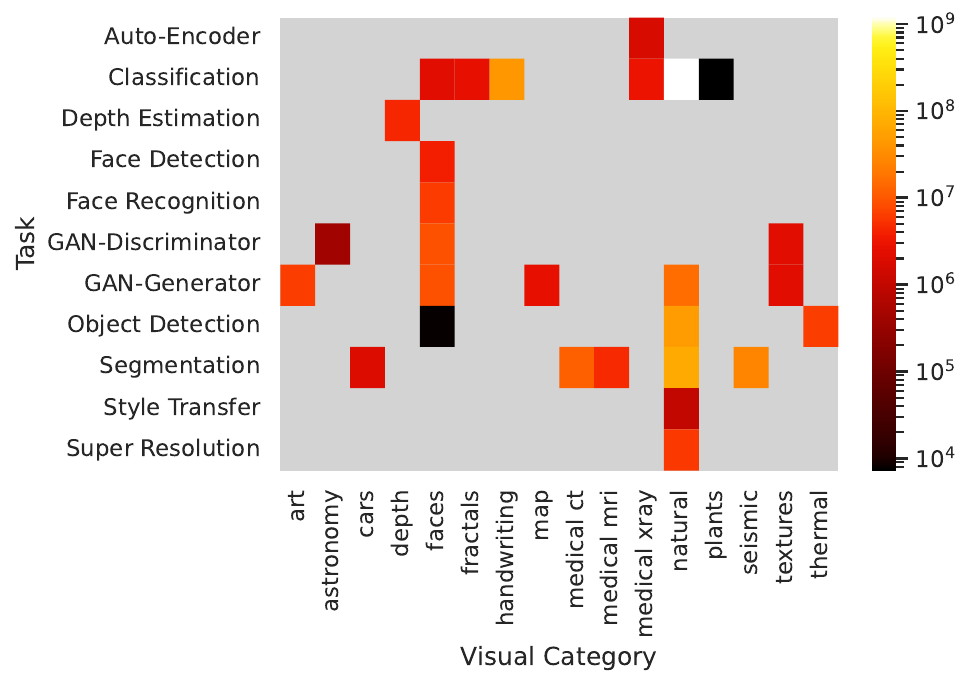}
    \caption{filter frequency}
  \end{subfigure}
  
  \caption{Bi-variate heatmap showing frequency aggregated by task and visual category.}
  \label{fig:stats_heat}
\end{figure*}

\section{Dataset details}

We provide \textit{CNN Filter DB} as a ca. 100 GB large HDF5 file which contains the unprocessed $3\times 3$ filters along with meta information as reported in \cref{tab:meta_columns}.  

We have collected models of the following tasks: \textit{Classification, GAN-Generator, Segmentation, Object Detection, Style Transfer, Depth Estimation, Face Detection, Super Resolution, GAN-Discriminator, Face Recognition, Auto-Encoder}.
The training sets were distributed into the following categories:  \textit{plants, natural, art, map, handwriting, medical ct, medical mri, depth, faces, textures, fractals, seismic, astronomy, thermal, medical xray, cars}.

\begin{figure*}
  \centering
  
  \begin{subfigure}{\linewidth}
    \includegraphics[width=\linewidth]{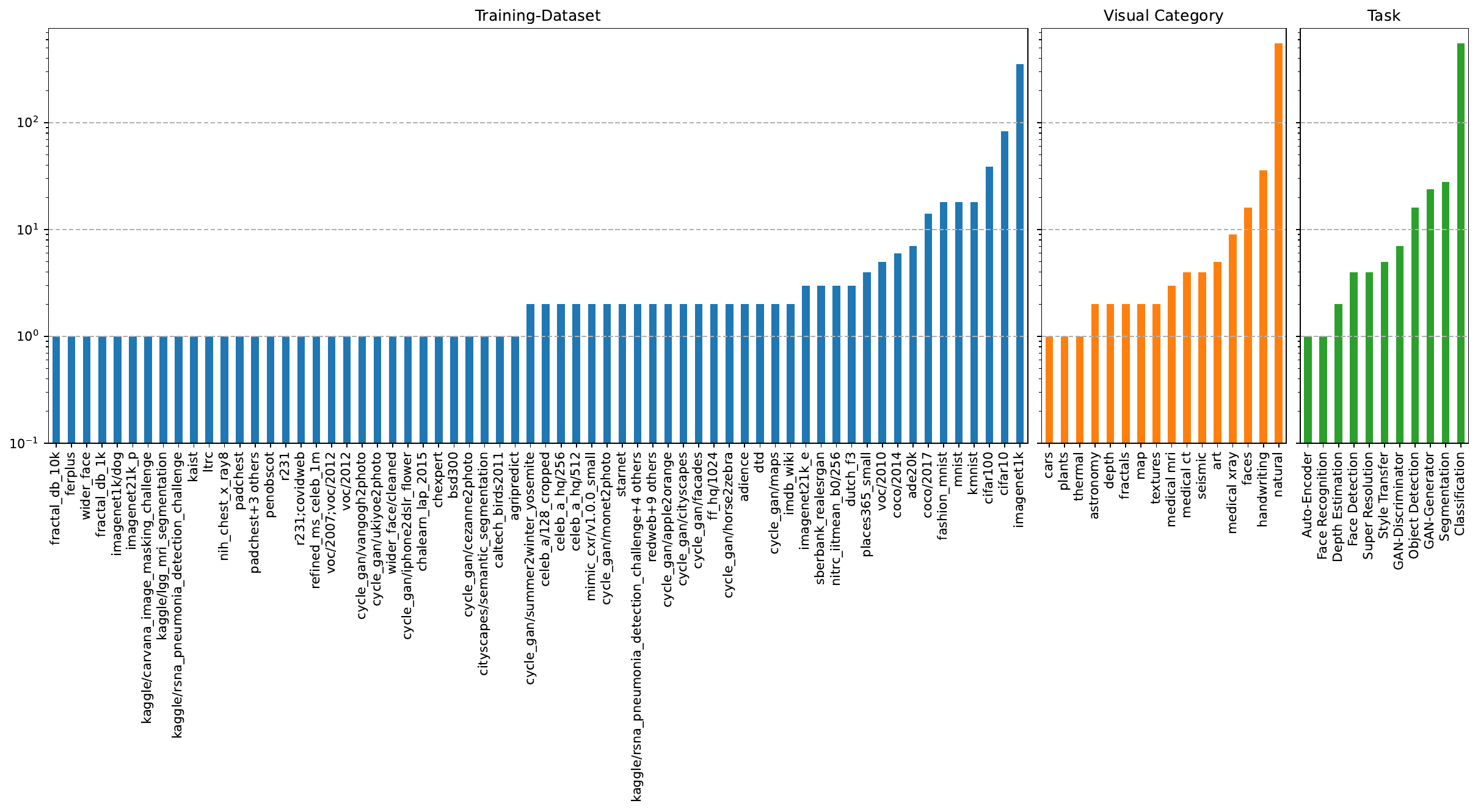}
    \caption{model frequency}
  \end{subfigure}
  
  \begin{subfigure}{\linewidth}
    \includegraphics[width=\linewidth]{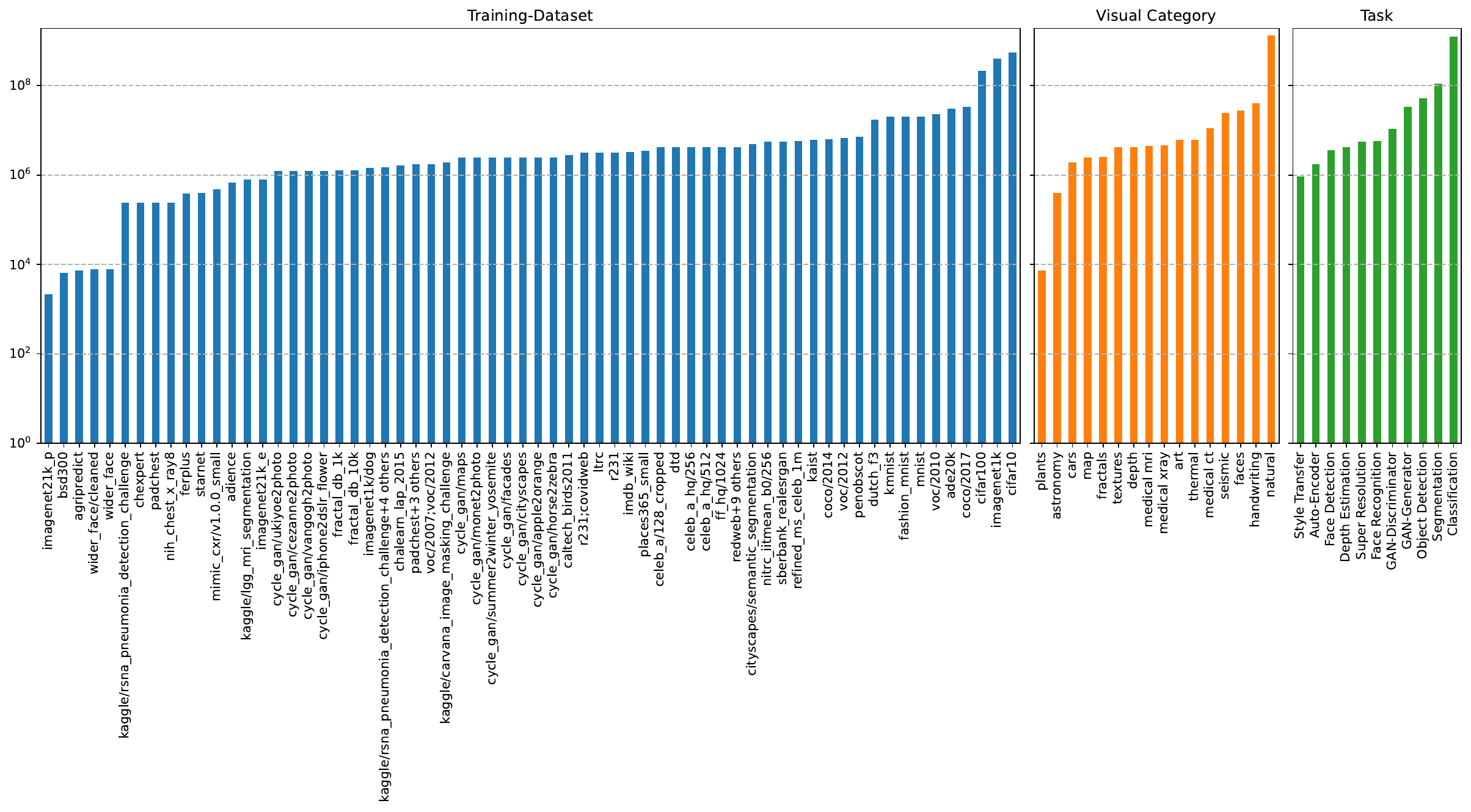}
    \caption{filter frequency}
  \end{subfigure}
  
  \caption{Total frequency per filter sub-set. Log scale.}
  \label{fig:stats}
\end{figure*}

A visualization of the accumulated frequency of models and filters by task, visual category, and training dataset combination can be found in \cref{fig:stats}. Heatmaps for aggregated frequency of filters/models by task and visual category are shown in \cref{fig:stats_heat}.

\begin{figure}
  \centering
  \includegraphics[width=\columnwidth]{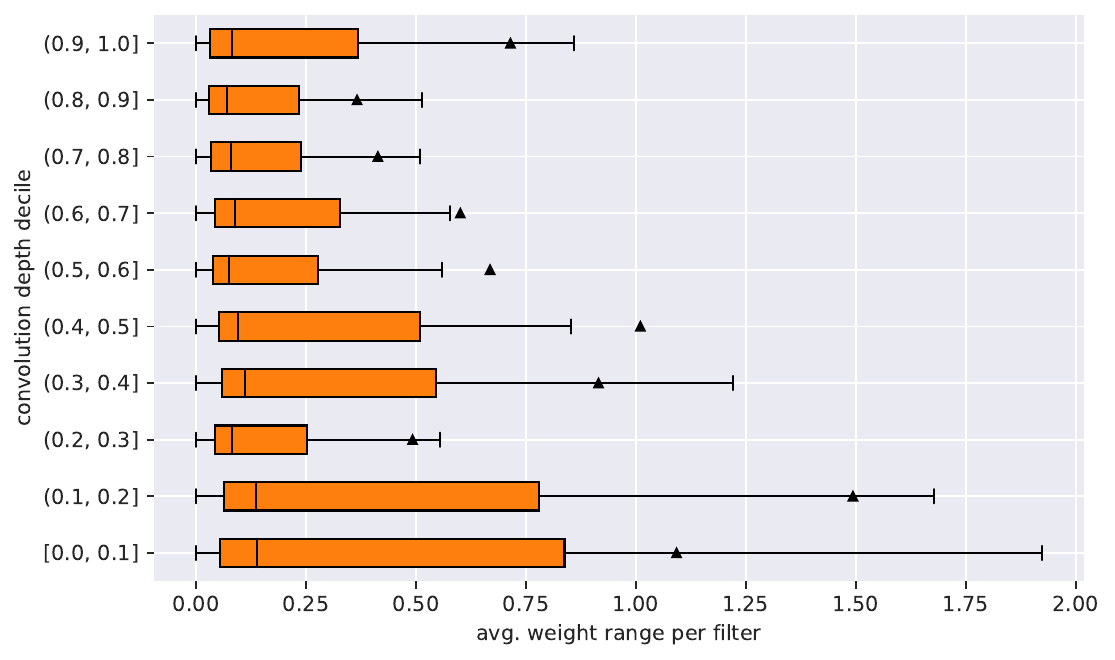}
  \caption{Boxplots showing average filter range per convolution depth decile (top to bottom in decreasing order) for each classification model in the dataset. Outliers are hidden for clarity.}
  \label{fig:range_boxplot}
\end{figure}

As previously mentioned, we used rescaled filters for all distribution shift related experiments. In \cref{fig:range_boxplot} we show the mean scale per layer depth decile of the unprocessed filters. We group the filters $f$ by model and depth decile in sets $S$ and compute the mean scale as follows:

\begin{equation}
    \begin{split}
        \hat{scale} = \sum_{f \in S} \frac{\max_{ij} f_{ij} - \min_{ij} f_{ij}}{|S|}
    \end{split}
\end{equation}

The distributions show an unsurprising decrease with depth but also a high variance and many outliers across models, especially in the first two deciles. 

Lastly, \cref{tab:models} contains all models we have used for our analysis.

\section{Derivation of randomness threshold}

\begin{figure}
    \includegraphics[width=\columnwidth]{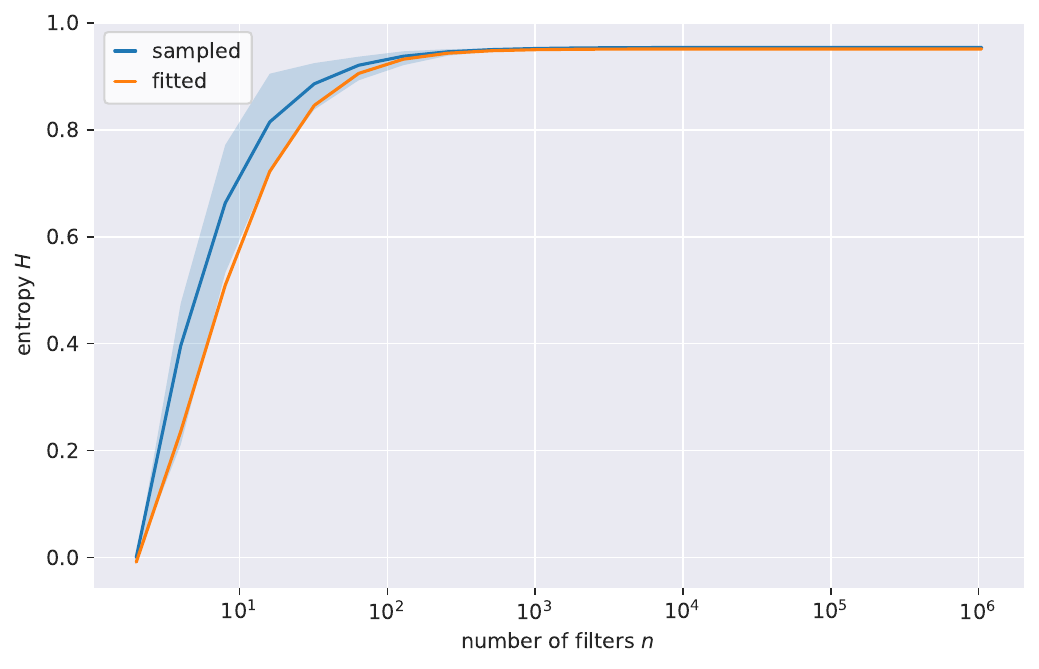}
    \caption{Sampled entropy for randomly initialized convolution layers with $n$ filters and fitted sigmoid $T_H$.}
    \label{fig:entropy_bound}
\end{figure}

We draw $n = {2^{1}, \dotsc, 2^{21}}$ filters with $3\times 3$ shape from a standard normal distribution and calculate the entropy $H$ as defined in the \textit{Methods} section. We repeat this process 1000 times for each $n$ and fit a sigmoid to the lowest entropy we have observed for each $n$. \cref{fig:entropy_bound} shows the obtained samples alongside the fitted sigmoid $T_H$.

\section{Ablation study: Degeneration impact}

\begin{figure}
  \centering
  \includegraphics[width=\columnwidth]{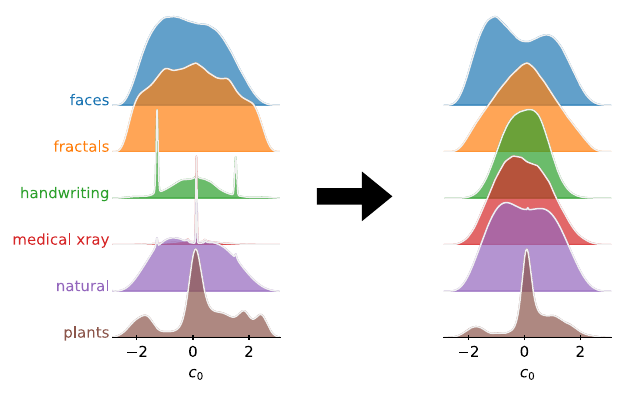}
  \caption{Most significant principal component KDEs of classification models by tasks before (left) and after removal of degenerated layers (right).}
  \label{fig:ridge_cleaning}
\end{figure}

\begin{figure*}

    \centering
    \begin{subfigure}{0.49\linewidth}
        \includegraphics[width=\linewidth]{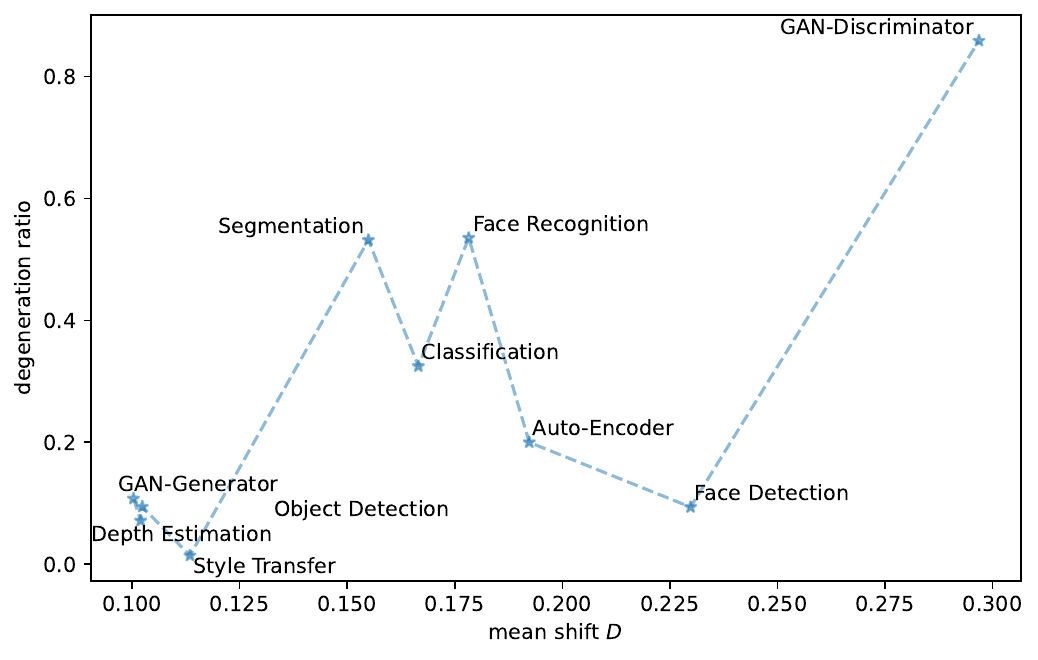}
        \caption{tasks}
    \end{subfigure}
    \begin{subfigure}{0.49\linewidth}
        \includegraphics[width=\linewidth]{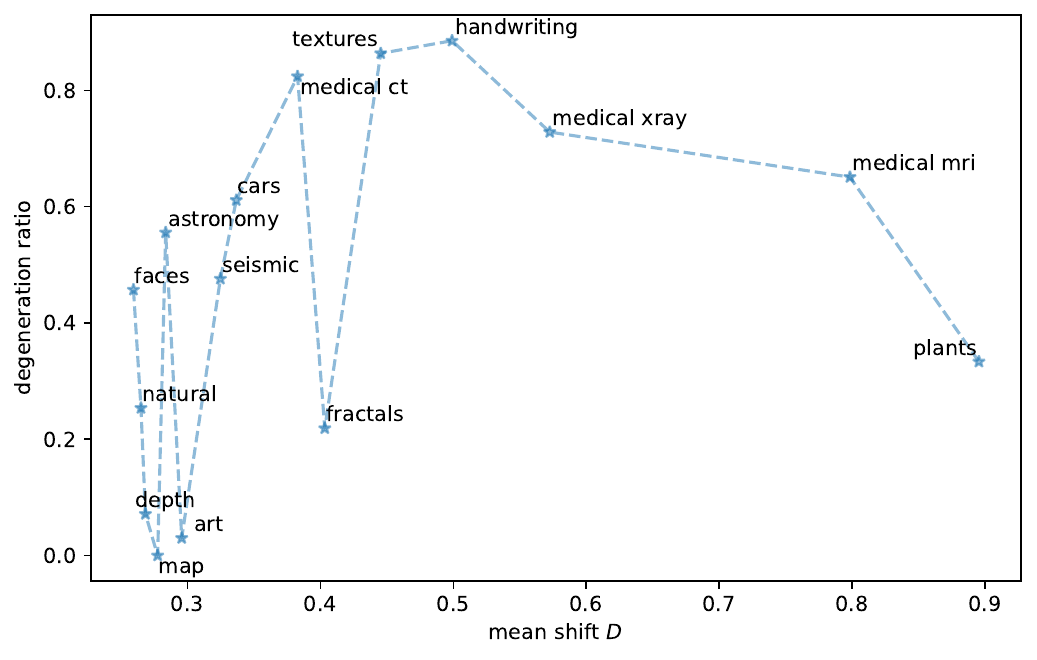}
        \caption{visual category}
    \end{subfigure}
    
    \caption{(Lack of) correlation between mean shift $D$ and layer degeneration ratio.}
    \label{fig:cleaning_impact}
\end{figure*}

\begin{figure*}
  \centering
  \includegraphics[width=\linewidth]{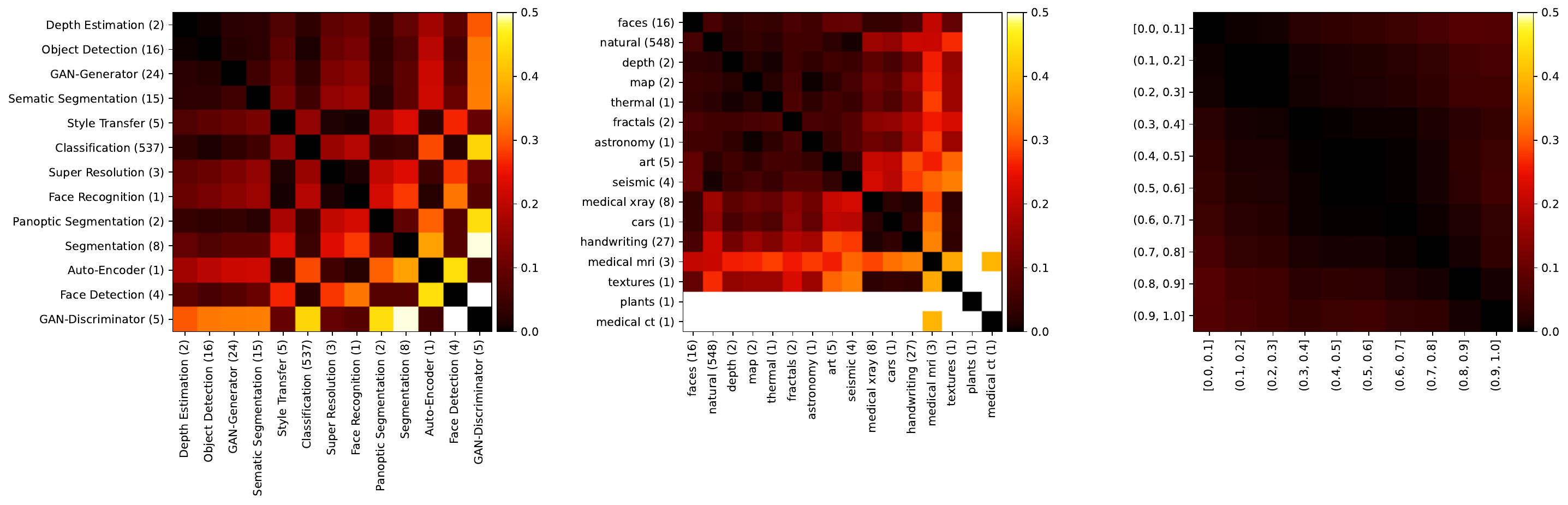}
  \begin{tabularx}{\linewidth}{>{\centering\arraybackslash}X>{\centering\arraybackslash}X>{\centering\arraybackslash}X}
        (a) tasks & (b) visual categories & (c) convolution depth decile\\
  \end{tabularx}
  \caption{Heatmaps over the shift $D$ for different filters groupings computed on the dataset \textit{without degenerated layers}. The number in brackets denotes the number of models in this group. Low values/dark colors denote low shifts.}
  \label{fig:kl_combined_clean}
\end{figure*}

As mentioned in the \textit{Limitations} section, we attempted to reproduce our experiments with a dataset that did not include filters from degenerated layers. We applied the following selection criterion to detect degeneration based on entropy $H$ and sparsity $S$ as defined in our \textit{Methods} section:
\begin{equation}
    \begin{split}
        (H \geq T_{H} - 0.02) \lor (H < 0.5) \land (S \geq 0.14)
    \end{split}
\end{equation}
While we had a solid foundation for the entropy upper bound (minus some noise), the lower bound for entropy and the bound for sparsity are based on the average we found in our datasets. Note that increasing the lower bound for $H$ results in more similar distributions and therefore lower shift. Hence, this value should be picked very carefully to not filter out vital layers. Sparsity is usually seen in peaks around the center of the KDEs. Tuning this value has a significant impact on the shift (\cref{fig:kl_combined_clean}) since the large center peaks increase the KL-Divergence significantly (\cref{fig:ridge_cleaning}). With the selected threshold we fail to find a meaningful correlation between the ratio of degenerated layers and the average shift to other groups (i.e. \textit{tasks} or \textit{visual categories}; \cref{fig:cleaning_impact}).

\section{Distribution shift by precision}

\begin{figure}
    \includegraphics[width=\columnwidth]{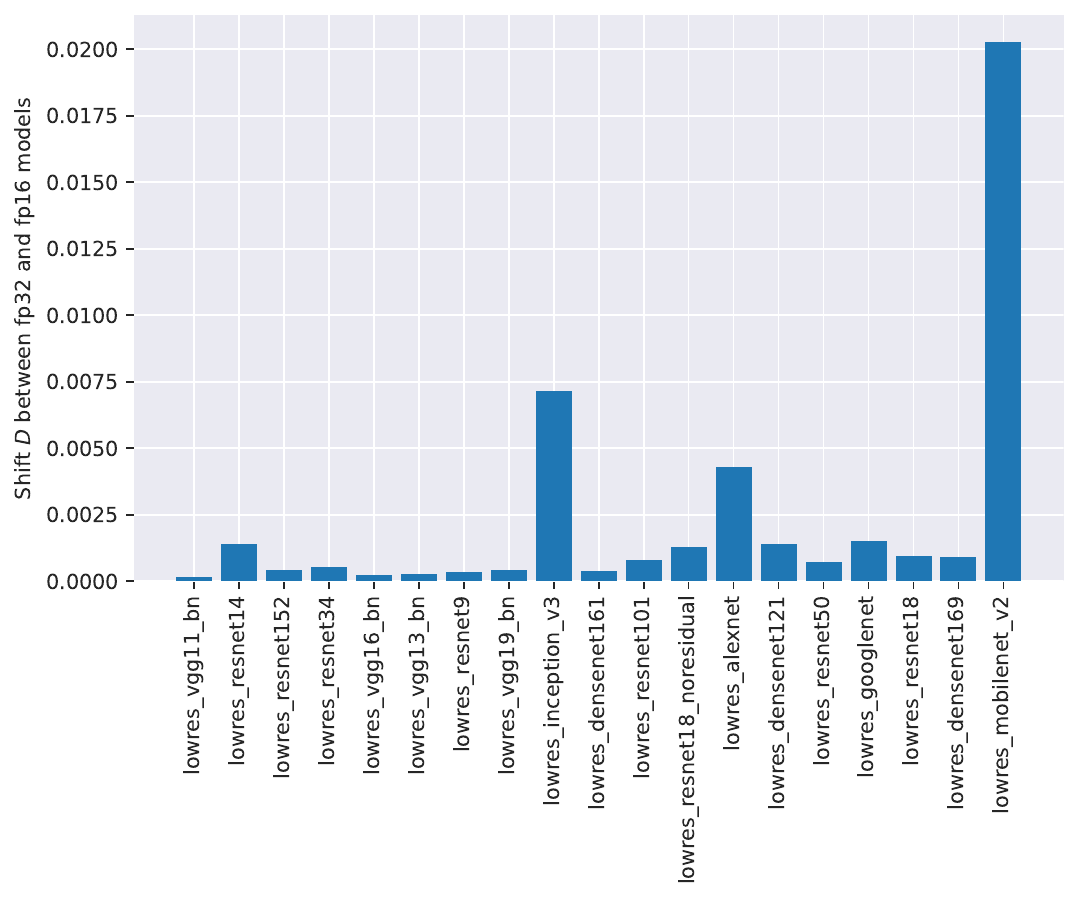}
    \caption{Distribution shift $D$ between low resolution models between trained on \textit{CIFAR-10} with fp16 and fp32 precision.}
    \label{fig:shift_precision}
\end{figure}

We initially assumed that quantization may lead to the \textit{spikes} phenotype, so we decided to test what shift we obtain when training with fp16 instead of fp32 precision. Spiky distributions should show high shifts in comparison to smooth distributions. We train all our low resolution models on \textit{CIFAR-10} with the same hyperparameters and observe marginal shifts \cref{fig:shift_precision}. Outliers with somewhat higher shifts include \textit{MobileNet\_v2}. But we have verified that the shift for \textit{MobileNet\_v2} does not exceed the shift one would measure by retraining with random seeds. The \textit{ResNet-9} shift does not exceed its retraining shift, therefore we assume that this also applies to other models.

\section{Distribution shift by convolution depth}

\begin{figure}
  \centering
  \includegraphics[width=\columnwidth]{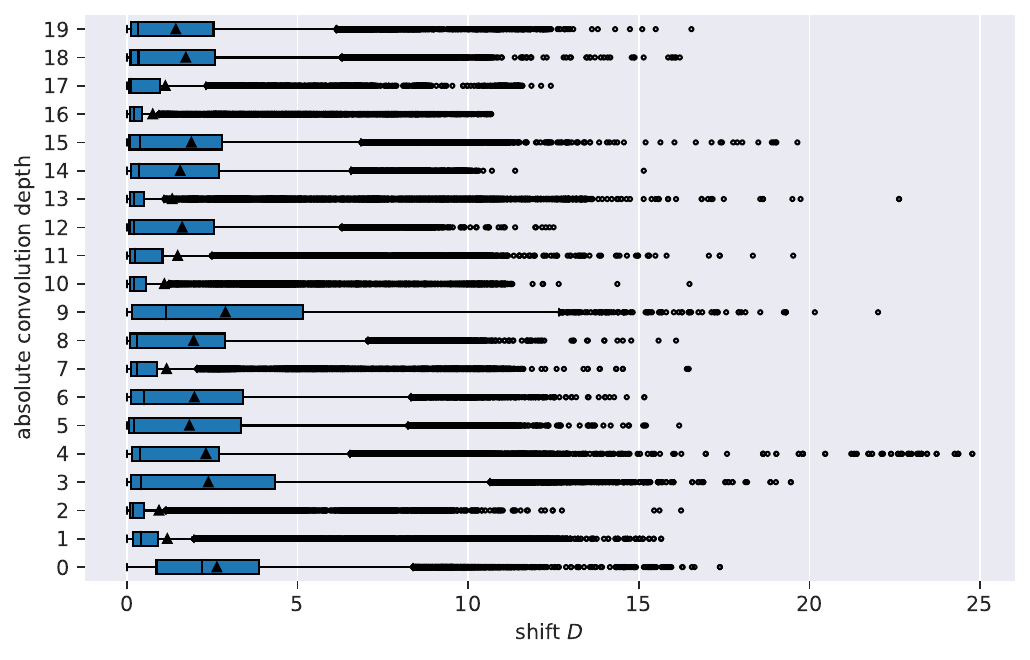}
  \caption{Boxplots showing the distribution of pair-wise model-to-model shift $D$ of \textit{classification} models per convolution depth. Our intentionally overparameterized models were left out of this analysis.}
  \label{fig:conv_depth_shift_abs}
\end{figure}

\begin{figure*}

    \centering
   \begin{subfigure}{0.49\linewidth}
        \includegraphics[width=\linewidth]{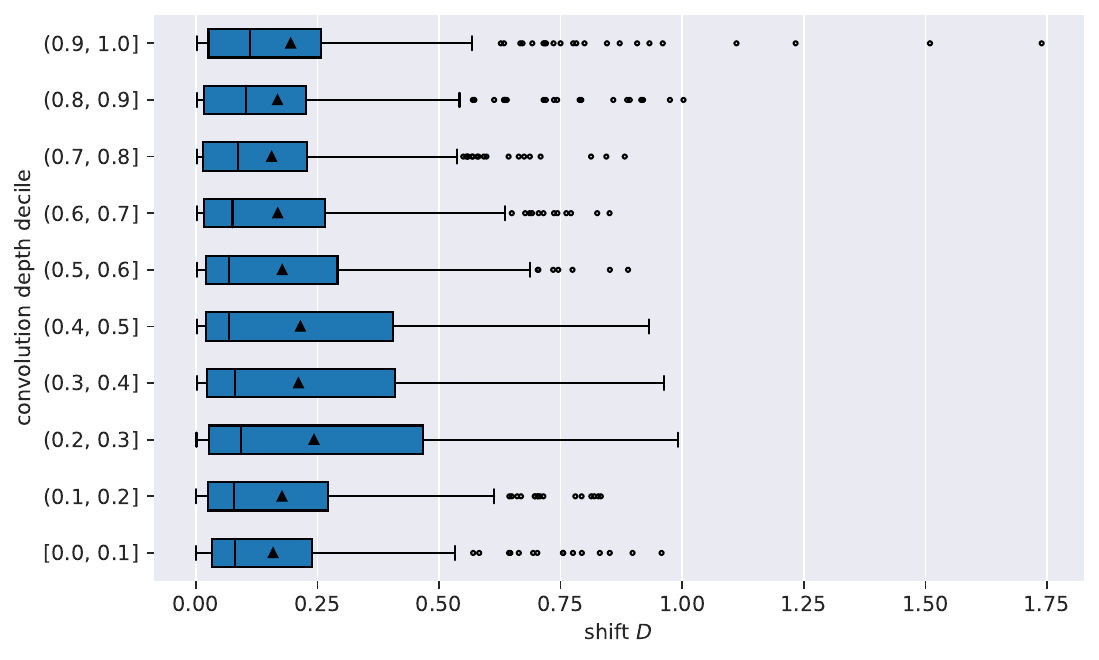}
        \caption{GAN-Generator}
    \end{subfigure}
    \begin{subfigure}{0.49\linewidth}
        \includegraphics[width=\linewidth]{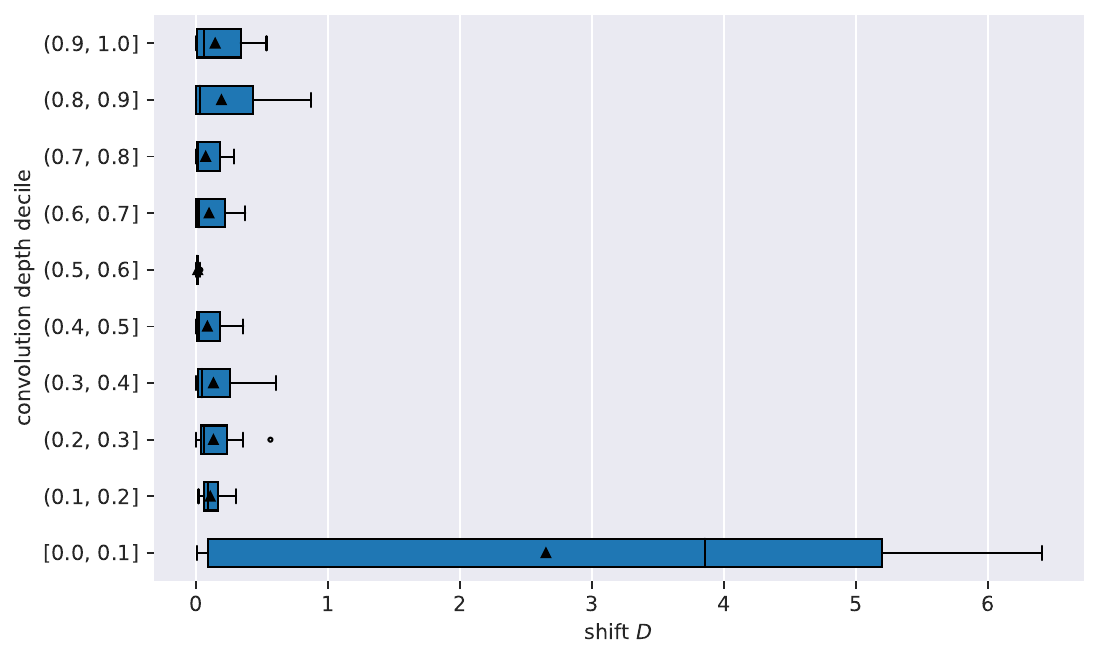}
        \caption{GAN-Discriminator}
    \end{subfigure}
    \begin{subfigure}{0.49\linewidth}
        \includegraphics[width=\linewidth]{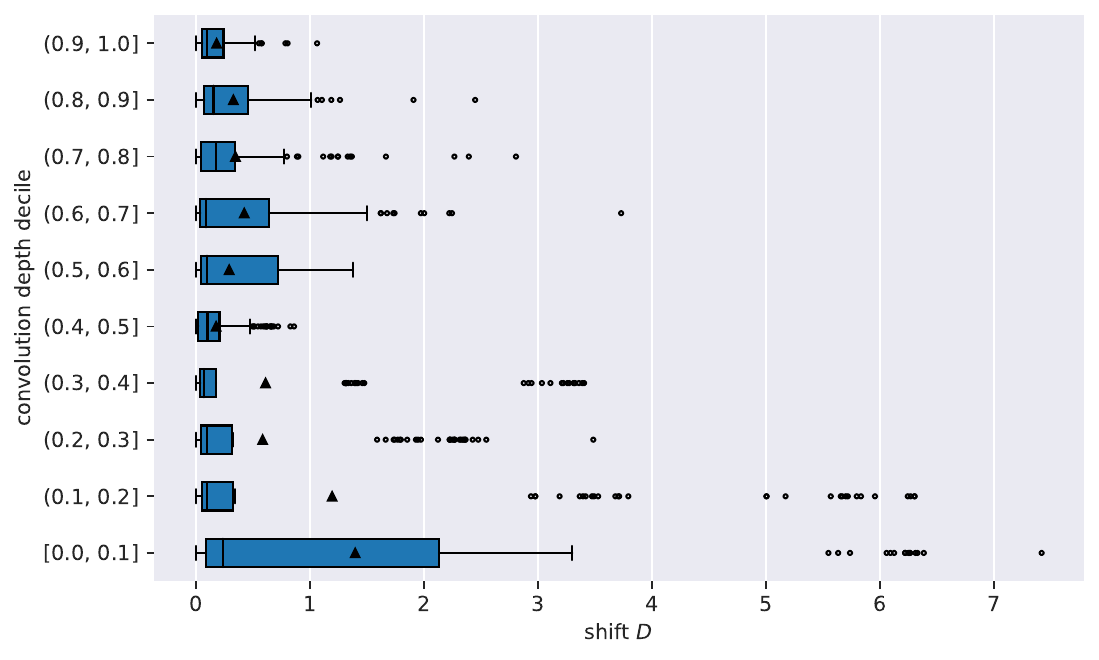}
        \caption{Object Detection}
    \end{subfigure}
    \begin{subfigure}{0.49\linewidth}
        \includegraphics[width=\linewidth]{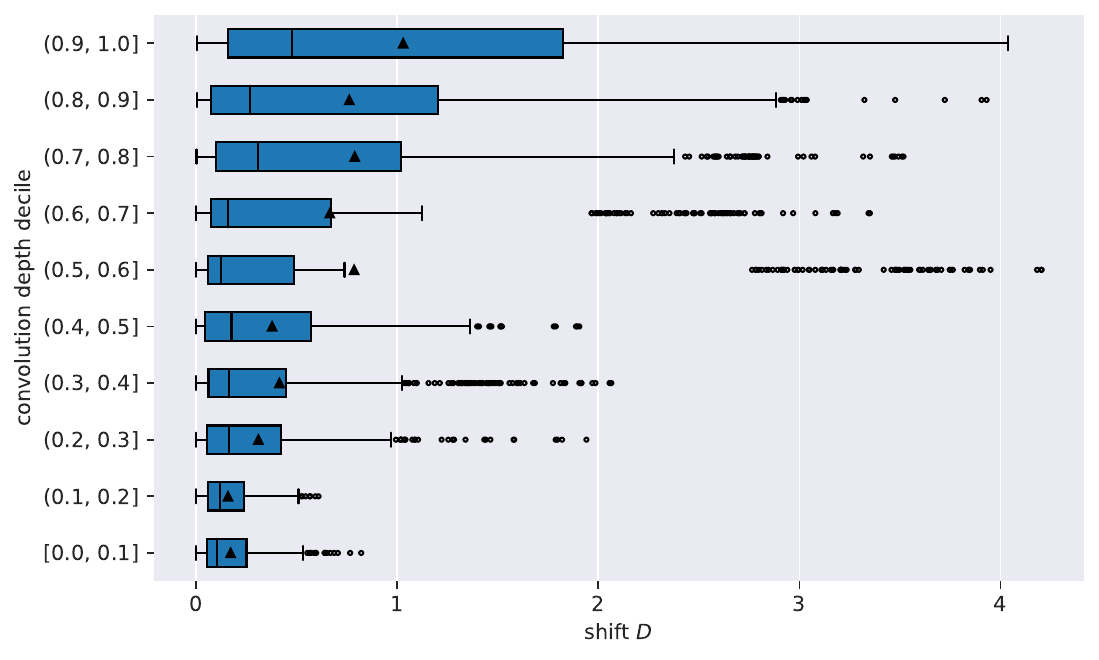}
        \caption{Segmentation}
    \end{subfigure}
    \begin{subfigure}{0.49\linewidth}
        \includegraphics[width=\linewidth]{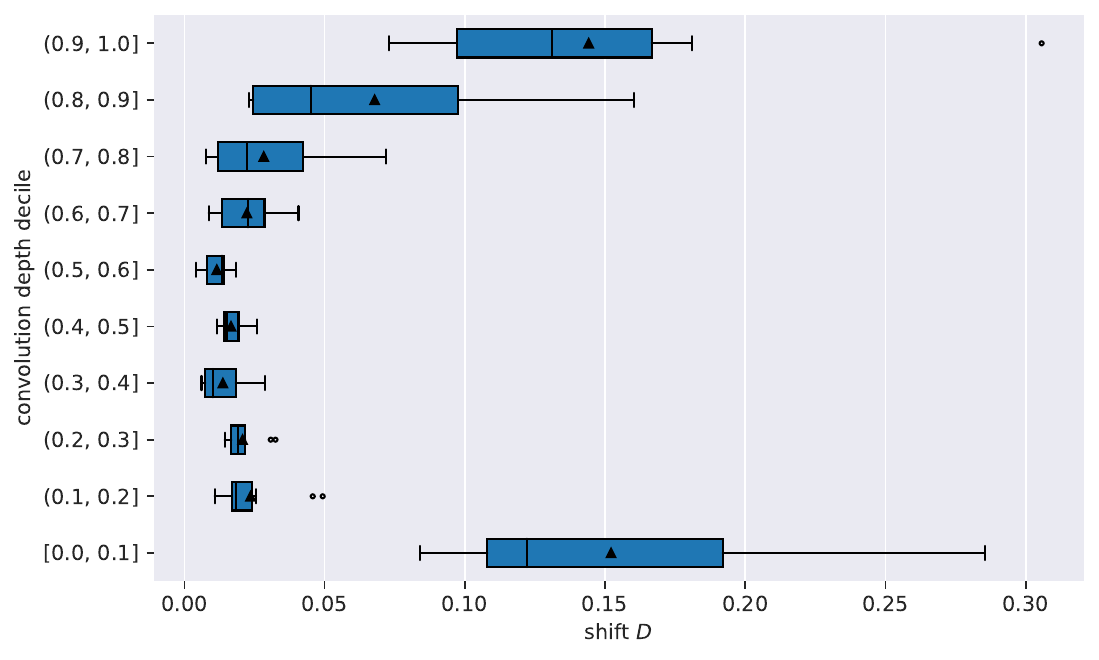}
        \caption{Style Transfer}
        \label{fig:conv_depth_shift_tasks_style}
    \end{subfigure}
    \begin{subfigure}{0.49\linewidth}
        \includegraphics[width=\linewidth]{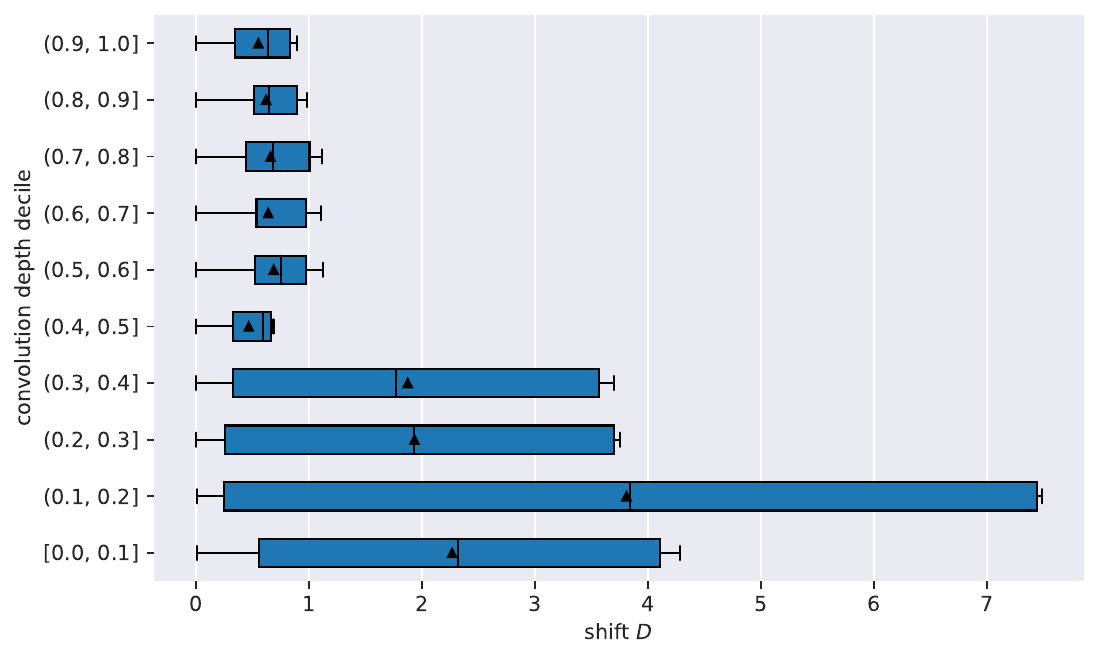}
        \caption{Face Detection}
    \end{subfigure}
    
    \caption{Boxplots showing the distribution of pair-wise model-to-model shift $D$ of models trained for various tasks per convolution depth decile.  Note the change in scale of the x-axis.}
    \label{fig:conv_depth_shift_tasks}
\end{figure*}

In addition to the main paper we also report the shift by absolute depth for the first 20 layers in \cref{fig:conv_depth_shift_abs} of \textit{classification} models and the shift by relative depth for more tasks in \cref{fig:conv_depth_shift_tasks}. Please note, that \cref{fig:conv_depth_shift_tasks_style} only contains the same network trained on different datasets. 

\section{More principal components}
\begin{figure*}

    \centering
    \begin{subfigure}{0.49\linewidth}
        \includegraphics[width=\linewidth]{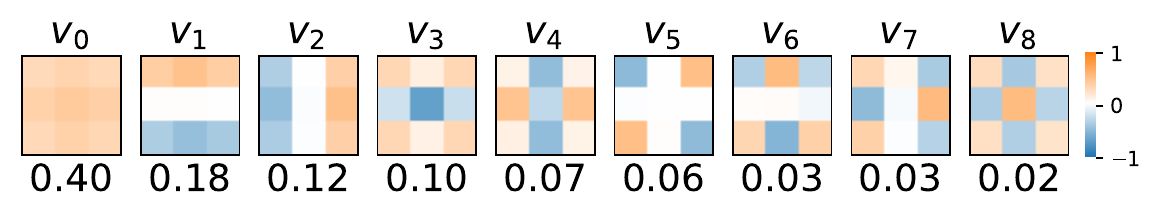}
        \caption{All filters}
    \end{subfigure}
    \begin{subfigure}{0.49\linewidth}
        \includegraphics[width=\linewidth]{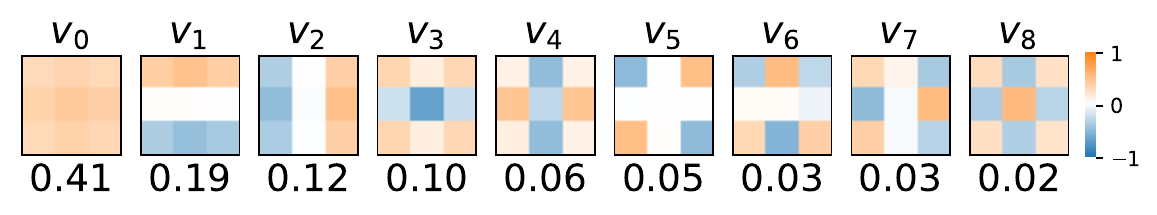}
        \caption{Task: classification}
    \end{subfigure}
    \begin{subfigure}{0.49\linewidth}
        \includegraphics[width=\linewidth]{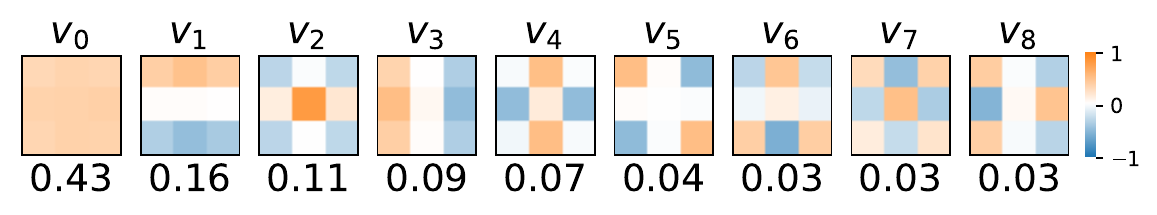}
        \caption{Dataset: ImageNet1k}
    \end{subfigure}
    \begin{subfigure}{0.49\linewidth}
        \includegraphics[width=\linewidth]{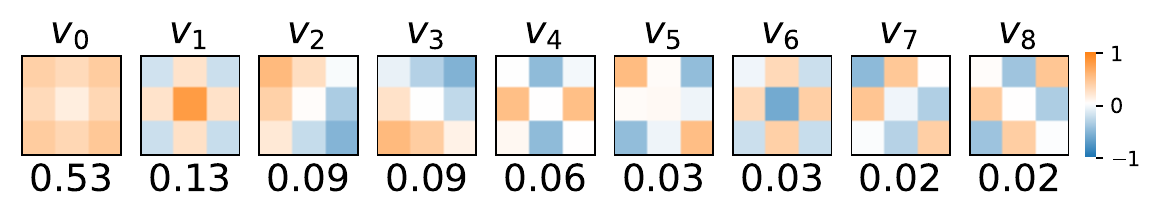}
        \caption{Visual Category: Fractals}
    \end{subfigure}
    \begin{subfigure}{0.49\linewidth}
        \includegraphics[width=\linewidth]{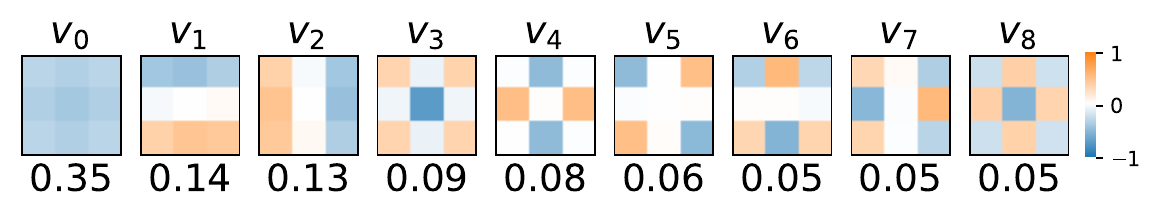}
        \caption{Task: GAN-Generator}
    \end{subfigure}
    \begin{subfigure}{0.49\linewidth}
        \includegraphics[width=\linewidth]{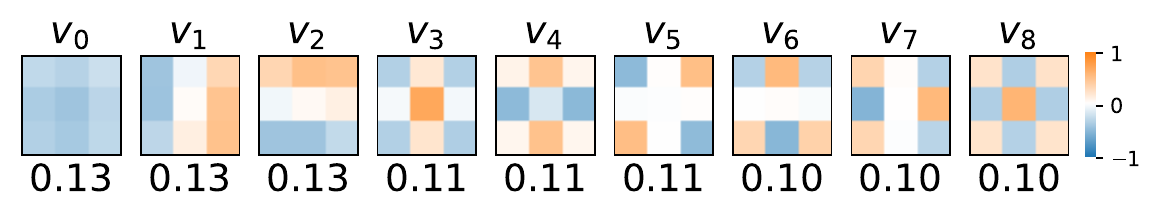}     
        \caption{Task: GAN-Discriminator}
    \end{subfigure}
    \begin{subfigure}{0.49\linewidth}
        \includegraphics[width=\linewidth]{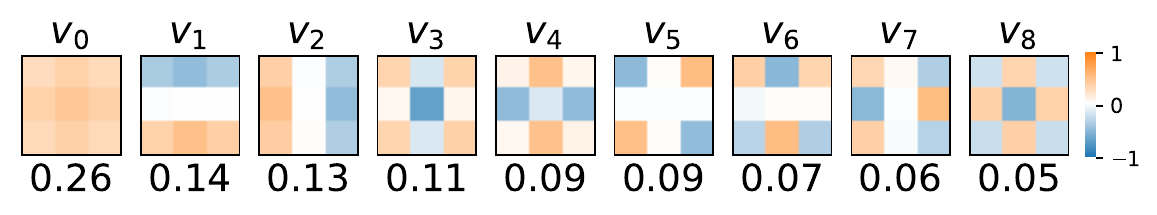}     
        \caption{First convolution layer}
    \end{subfigure}
    \begin{subfigure}{0.49\linewidth}
        \includegraphics[width=\linewidth]{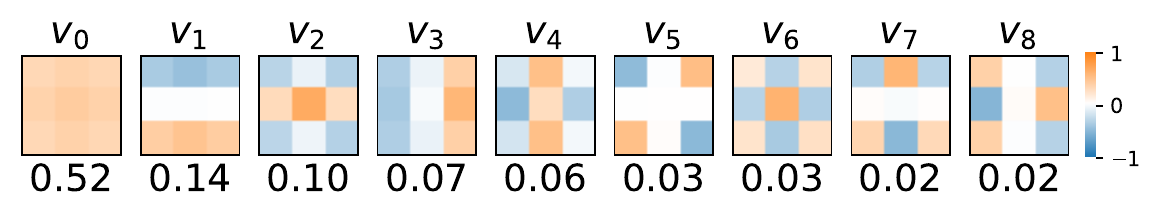} 
        \caption{Last convolution layers}
    \end{subfigure}
    
    \caption{Depiction of the filter basis and (cumulative) explained variance ratio per component for filters grouped by various meta-data dimensions.}
    \label{fig:appendix_filter_basis}
\end{figure*}

\begin{figure*}
    \centering
    
    \begin{subfigure}{0.49\linewidth}
        \includegraphics[width=\linewidth]{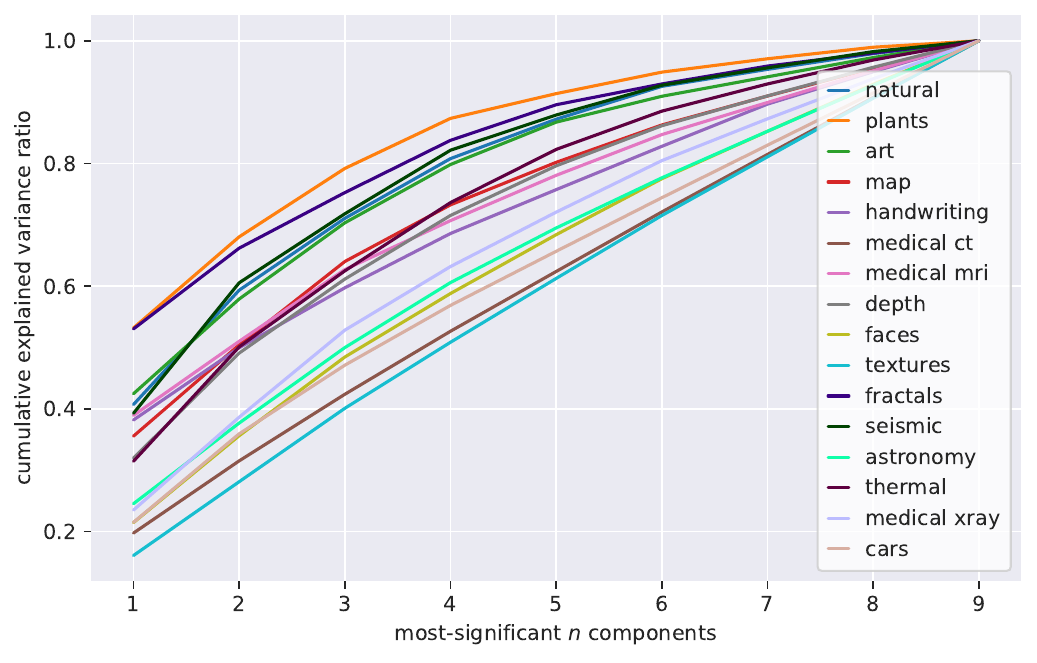}
        \caption{visual category}
    \end{subfigure}
    \begin{subfigure}{0.49\linewidth}
        \includegraphics[width=\linewidth]{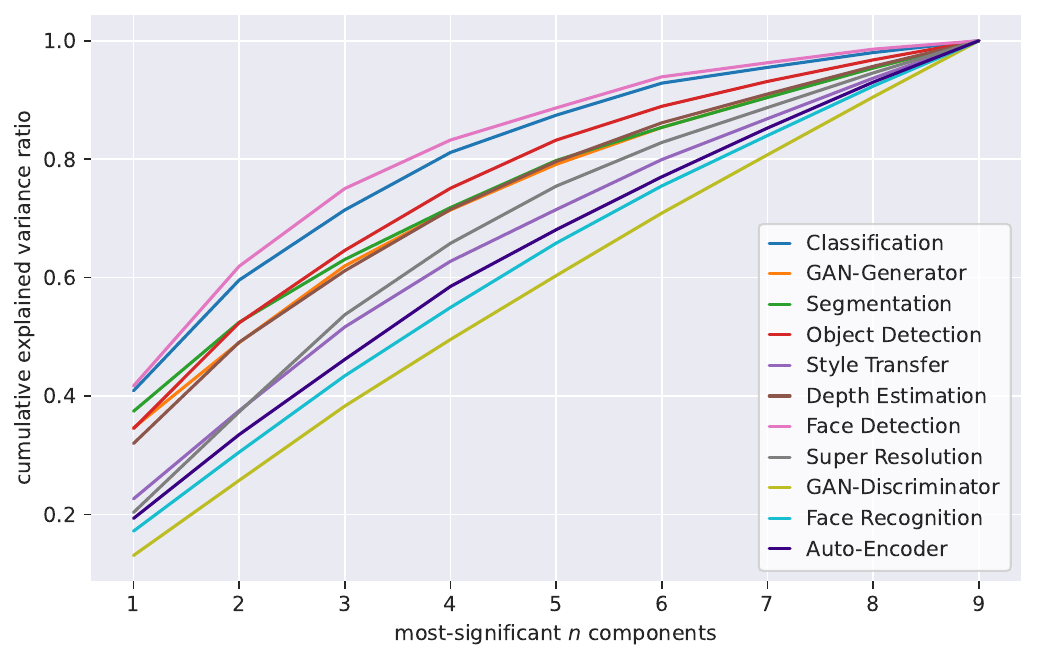}
        \caption{task}
    \end{subfigure}
    
    \caption{Cumulative ratio of explained variance over the first $n$ components by all tasks and visual categories.}
    \label{fig:appendix_pca_cumsum}
\end{figure*}

In \cref{fig:appendix_filter_basis} we add interesting counter-parts to the filter basis shown in the main paper. As one can observe the filter basis remains quite similar. Changes usually affect the order of the components (since they are sorted by explained variance ratio), inversion (though this is not characteristic, since the coefficients can simply be inverted), and noise presumably due to degeneration. \cref{fig:appendix_pca_cumsum} contains the cumulative explained variance ratio plots for all tasks and visual categories. 
For the sake of completeness, we also add that SVD centering $\displaystyle \bar{X} = [-0.04262863,\allowbreak -0.0411367,\allowbreak -0.04461834,\allowbreak -0.0407119,\allowbreak -0.03574134,\allowbreak
       -0.04268694,\allowbreak -0.04350573,\allowbreak -0.04138637,\allowbreak -0.04486743]$ for the full dataset.

\section{More KDEs plots}

\begin{figure*}
  \centering
  \includegraphics[width=\linewidth]{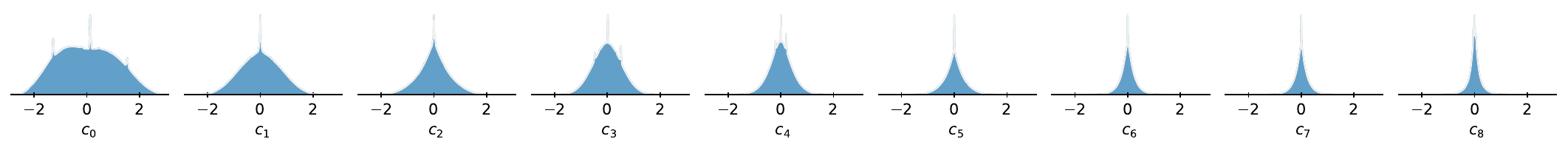}
  \caption{Distribution of the coefficients along the principal components of the \textbf{full dataset}.}
    \label{fig:ridge_basis}
\end{figure*}

\begin{figure*}
  \centering
  \includegraphics[width=\linewidth]{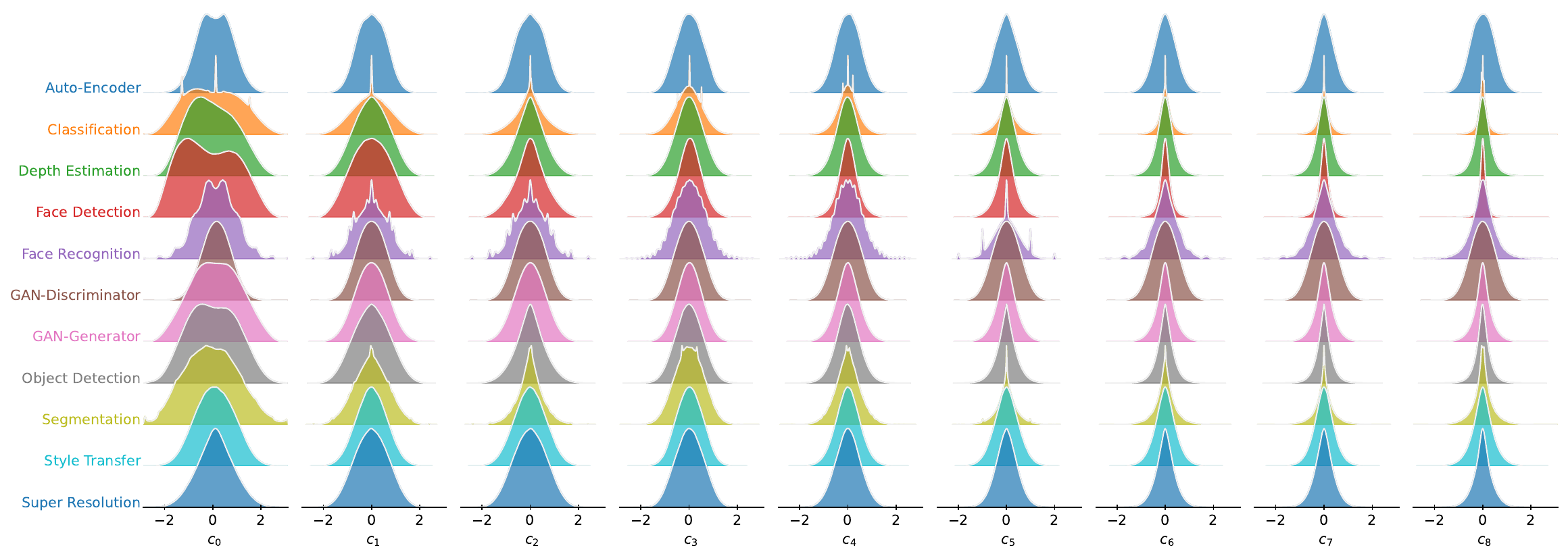}
  \caption{Distribution of the coefficients along the principal components by \textbf{model task}.}
    \label{fig:ridge_task}
\end{figure*}

\begin{figure*}
  \centering
  \includegraphics[width=\linewidth]{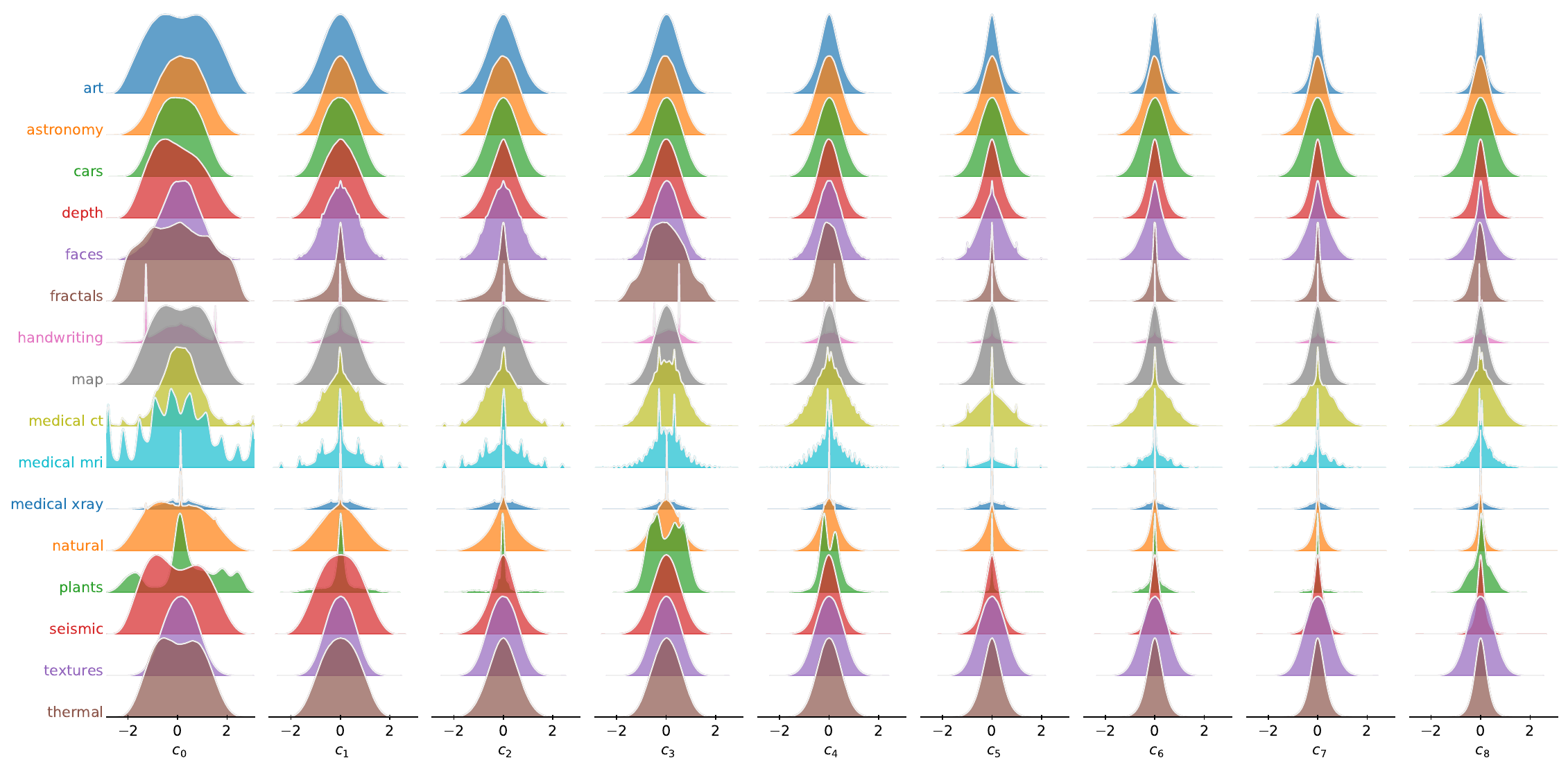}
  \caption{Distribution of the coefficients along the principal components by \textbf{visual category}.}
  \label{fig:ridge_visual_category}
\end{figure*}

\begin{figure*}
  \centering
  \includegraphics[width=\linewidth]{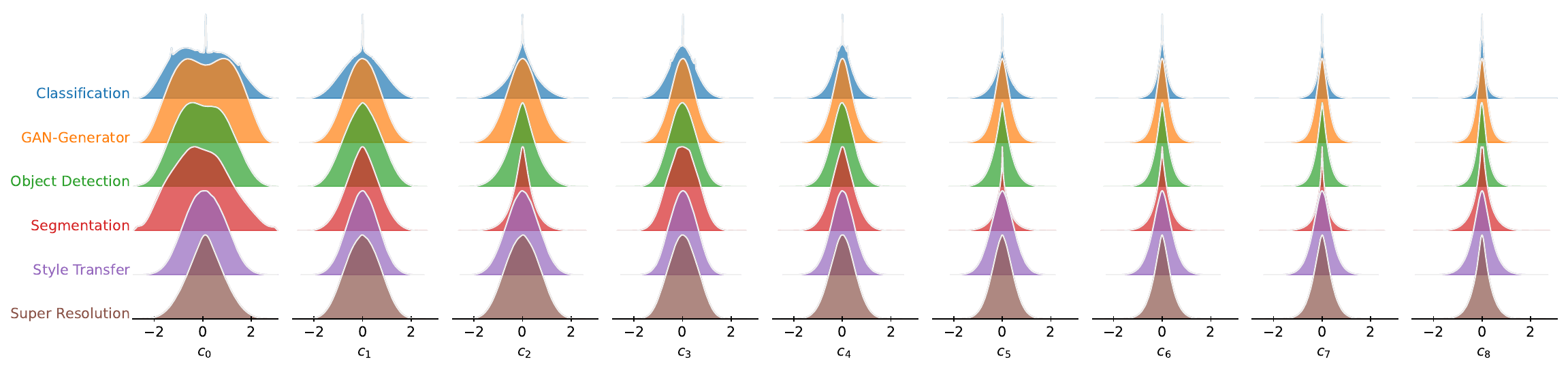}
  \caption{Distribution of the coefficients along the principal components by \textbf{model task for} datasets belonging to the \textbf{natural visual category.}}
  \label{fig:ridge_task_for_natural}
\end{figure*}

\begin{figure*}
  \centering
  \includegraphics[width=\linewidth]{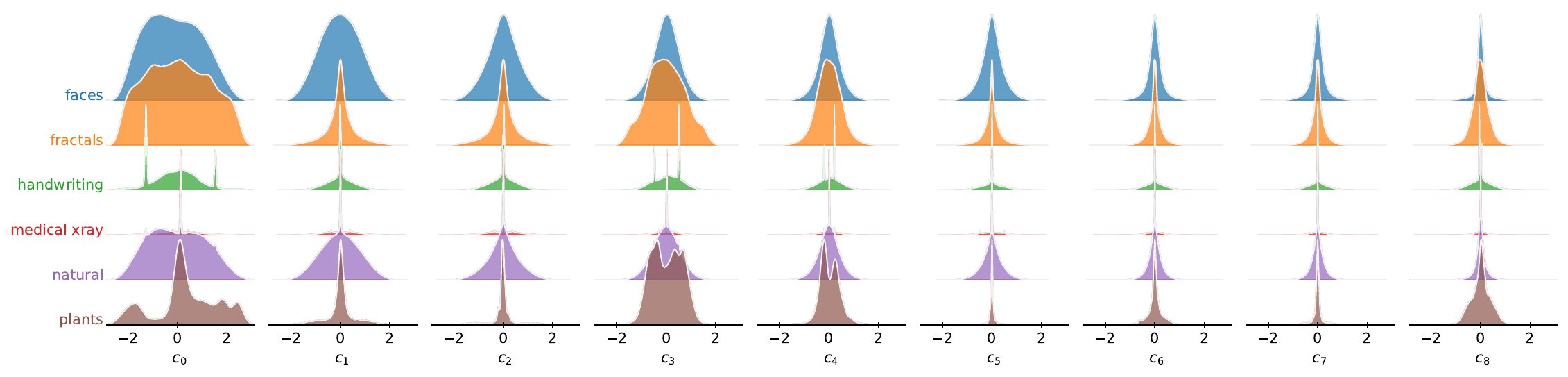}
  \caption{Distribution of the coefficients along the principal components by \textbf{visual training category for image classification models}.}
  \label{fig:ridge_visual_category_for_classification}
\end{figure*}

\begin{figure*}
  \centering
  \includegraphics[width=\linewidth]{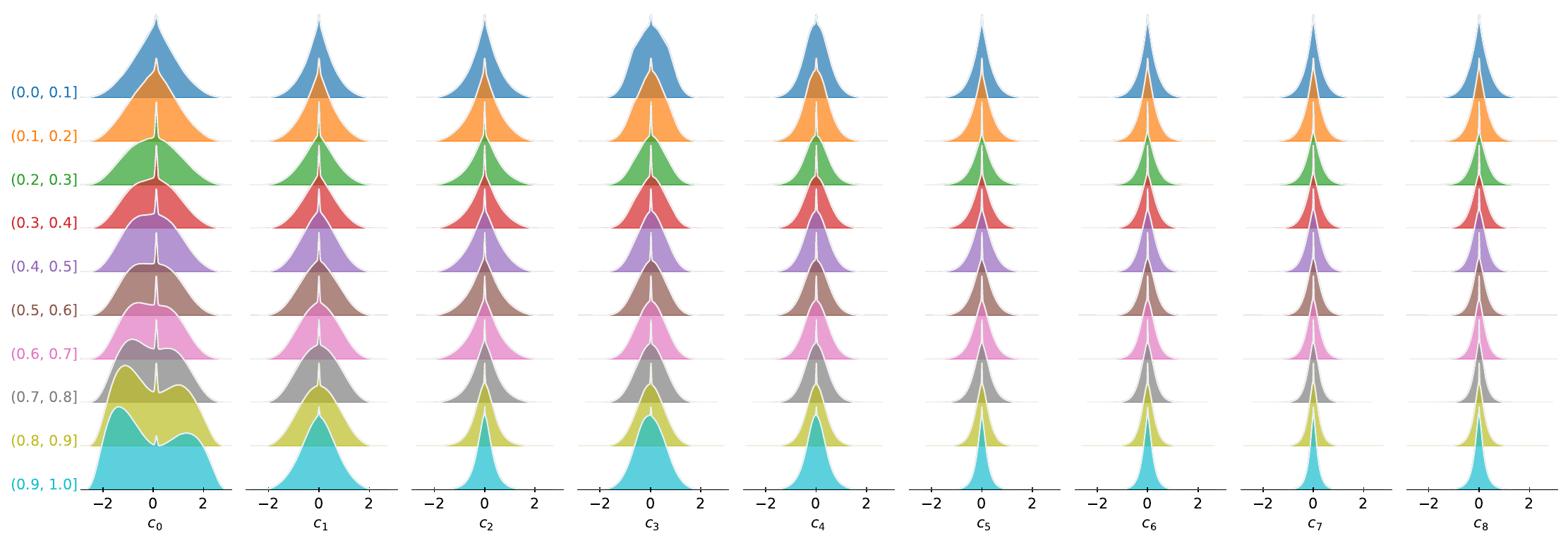}
  \caption{Distribution of the coefficients along the principal components by \textbf{convolution depth decile for image classification models}.}
  \label{fig:ridge_conv_depth_decile_classification}
\end{figure*}

\begin{figure*}
  \centering
  \includegraphics[width=\linewidth]{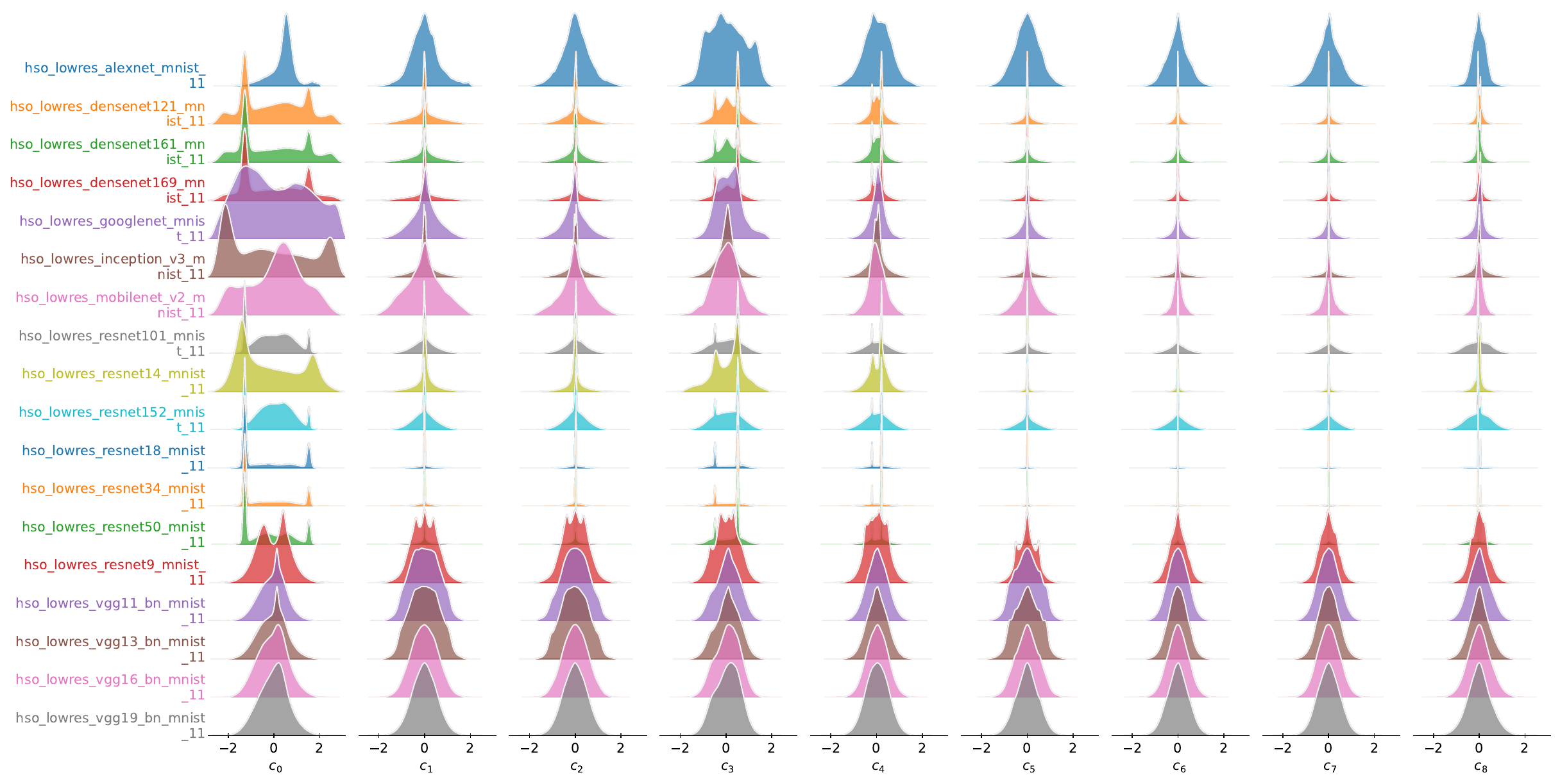}
  \caption{Distribution of the coefficients along the principal components of models trained on the \textbf{{MNIST}} dataset. All these models belong to our \textbf{intentionally overparameterized models}.}
  \label{fig:ridge_mnist_models}
\end{figure*}

\begin{figure*}
  \centering
  \includegraphics[width=\linewidth]{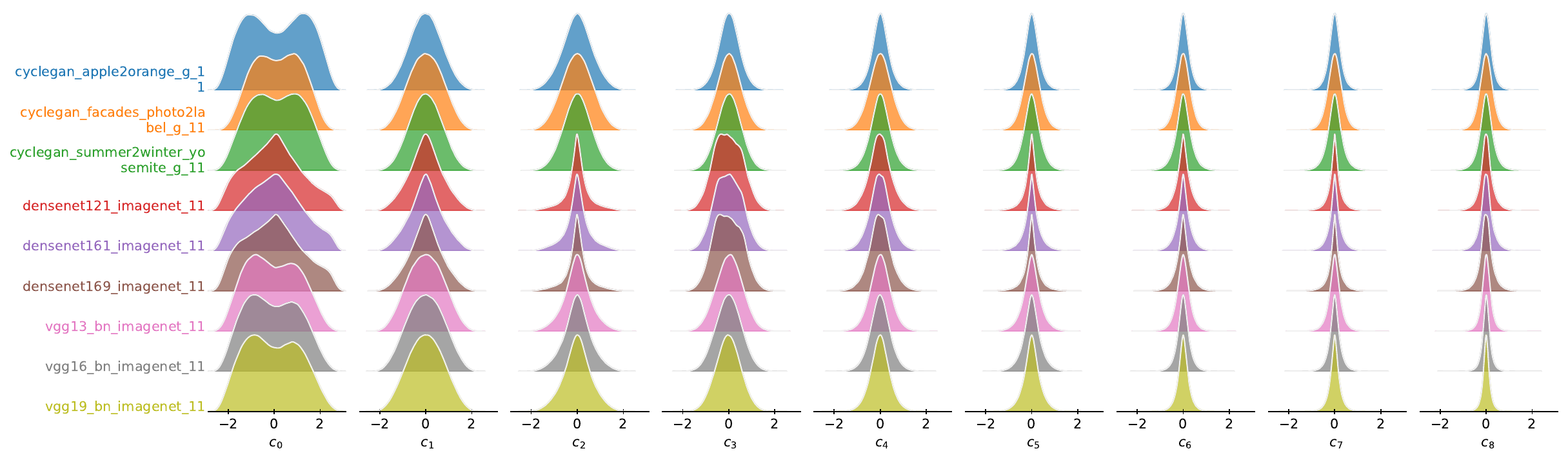}
  \caption{Distribution of the coefficients along the principal components of \textbf{selected models from similar families}.}
  \label{fig:ridge_model_family}
\end{figure*}

\cref{fig:ridge_basis} shows the KDEs (KDEs created with \cite{kdepy}) for every principal component on the full dataset. 
\cref{fig:ridge_task} shows KDEs for all tasks. 
\cref{fig:ridge_task_for_natural} shows only KDEs of tasks of models that were trained with datasets belonging to the natural visual category. 
\cref{fig:ridge_visual_category} shows KDEs for every visual category. Some categories show shifts due to bias representation while other clearly contain a majority of degenerated filters.
\cref{fig:ridge_visual_category_for_classification} show KDEs by the visual category of the training dataset limited to classification models. Several categories such as \textit{medical xray, plants, handwriting} are clearly impacted by degeneration.
\cref{fig:ridge_conv_depth_decile_classification} shows KDEs of classification models split by convolution depth decile. The distribution shift with depth reminds us of the shift of all filters we have seen during training \textit{ResNet-9} in our \textit{Results} section.
\cref{fig:ridge_model_family} shows some selected models from the same family, showing clear shifts between the families but low shifts within.
Lastly, \cref{fig:ridge_mnist_models} shows all models trained on \textit{MNIST}. These are consist exclusively of the intentionally overparameterized models. The KDEs show very clear signs of major degeneration, by stark spikes, especially around null.

\section{Phenotype scatter plots}

\begin{figure*}
    \centering
    \includegraphics[width=\linewidth]{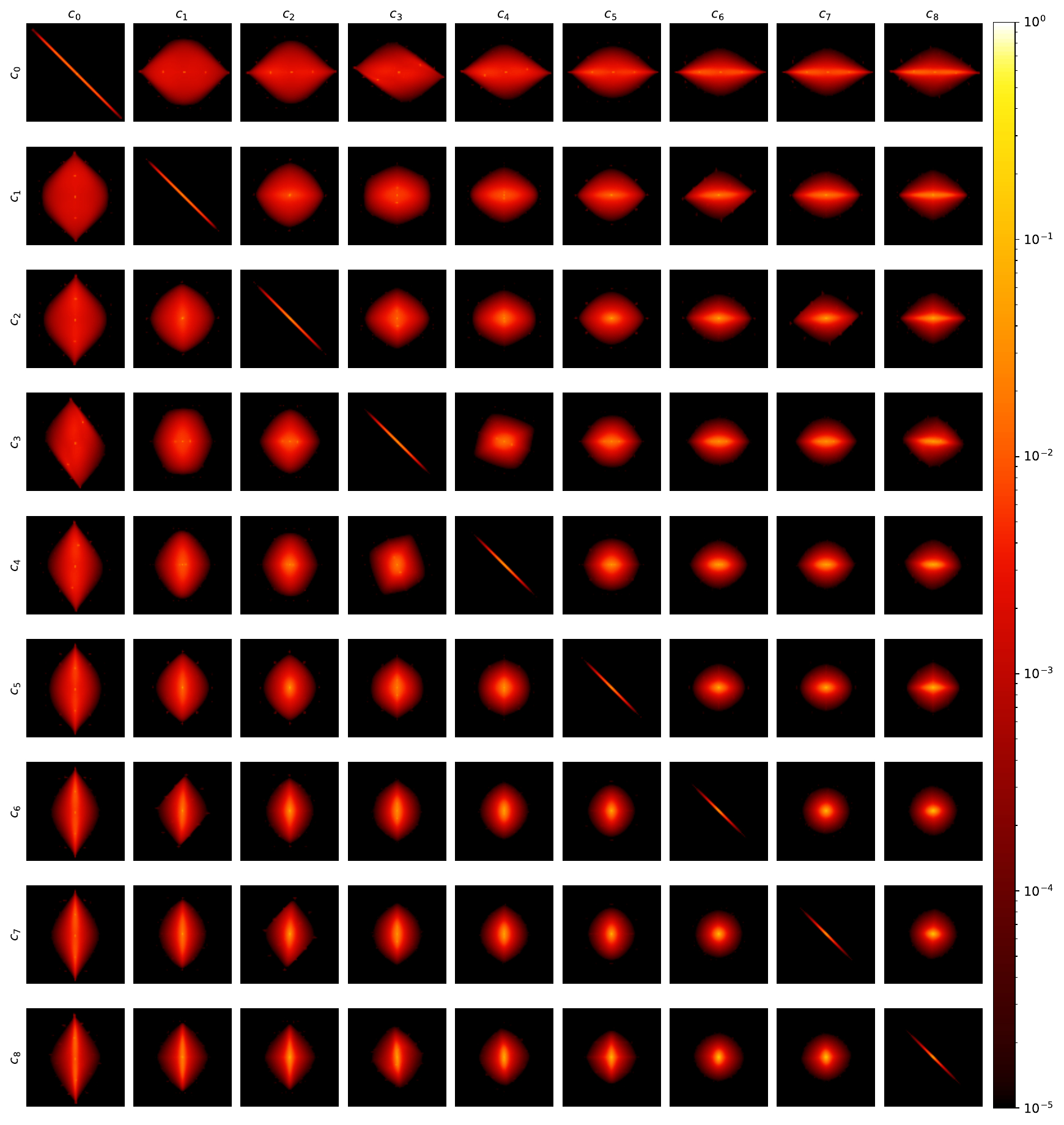}
    \caption{Bi-variate coefficient scatter plot over the full dataset.}
    \label{fig:pheno_full_all}
\end{figure*}

\begin{figure*}
    \centering
    \includegraphics[width=\linewidth]{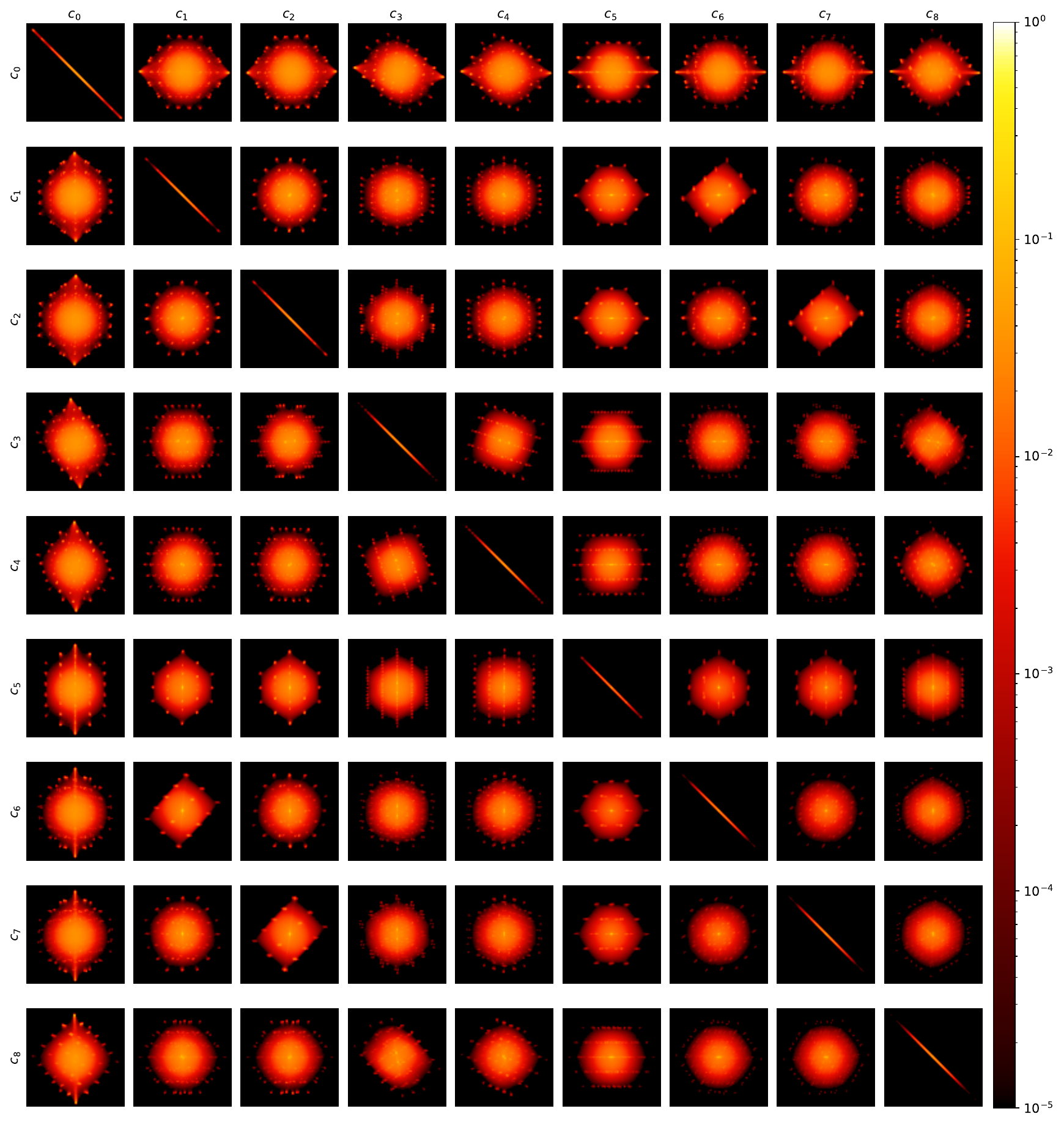}
    \caption{Bi-variate coefficient scatter plot of the phenotype \textit{spikes}.}
    \label{fig:pheno_full_spikes}
\end{figure*}

\begin{figure*}
    \centering
    \includegraphics[width=\linewidth]{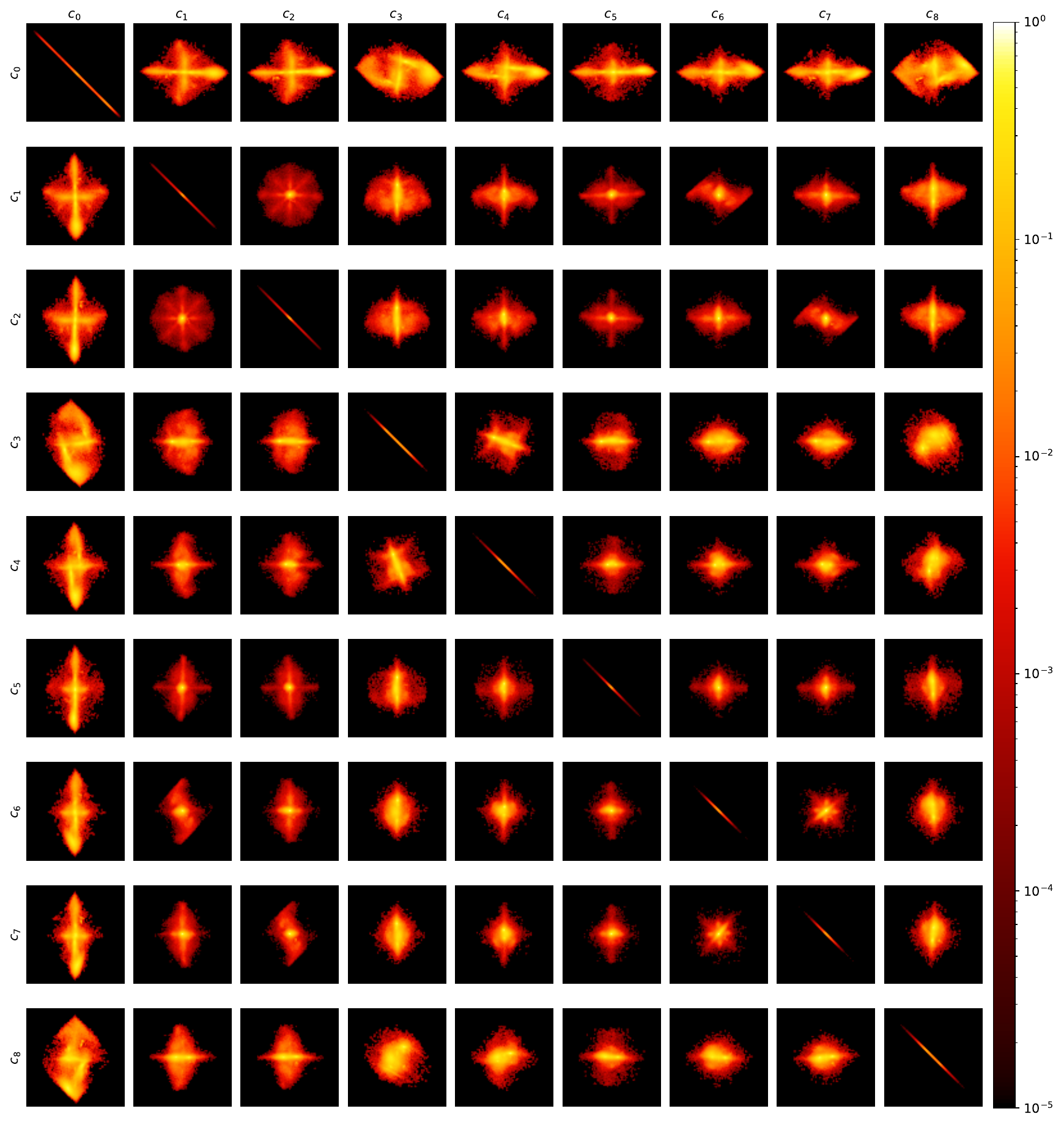}
    \caption{Bi-variate coefficient scatter plot of the phenotype \textit{symbols}.}
    \label{fig:pheno_full_symbols}
\end{figure*}

\begin{figure*}
    \centering
    \includegraphics[width=\linewidth]{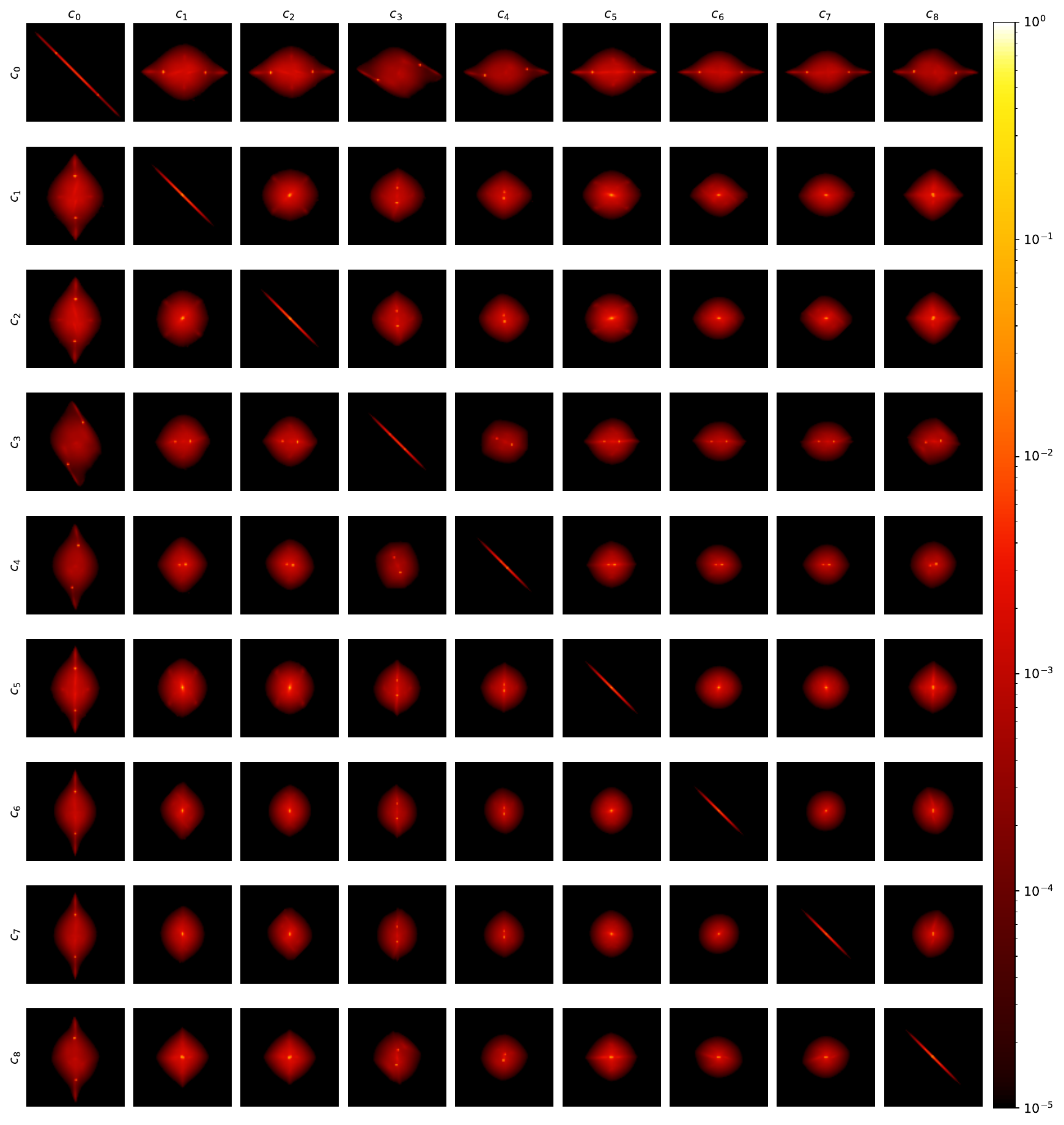}
    \caption{Bi-variate coefficient scatter plot of the phenotype \textit{point}.}
    \label{fig:pheno_full_point}
\end{figure*}

\begin{figure*}
    \centering
    \includegraphics[width=\linewidth]{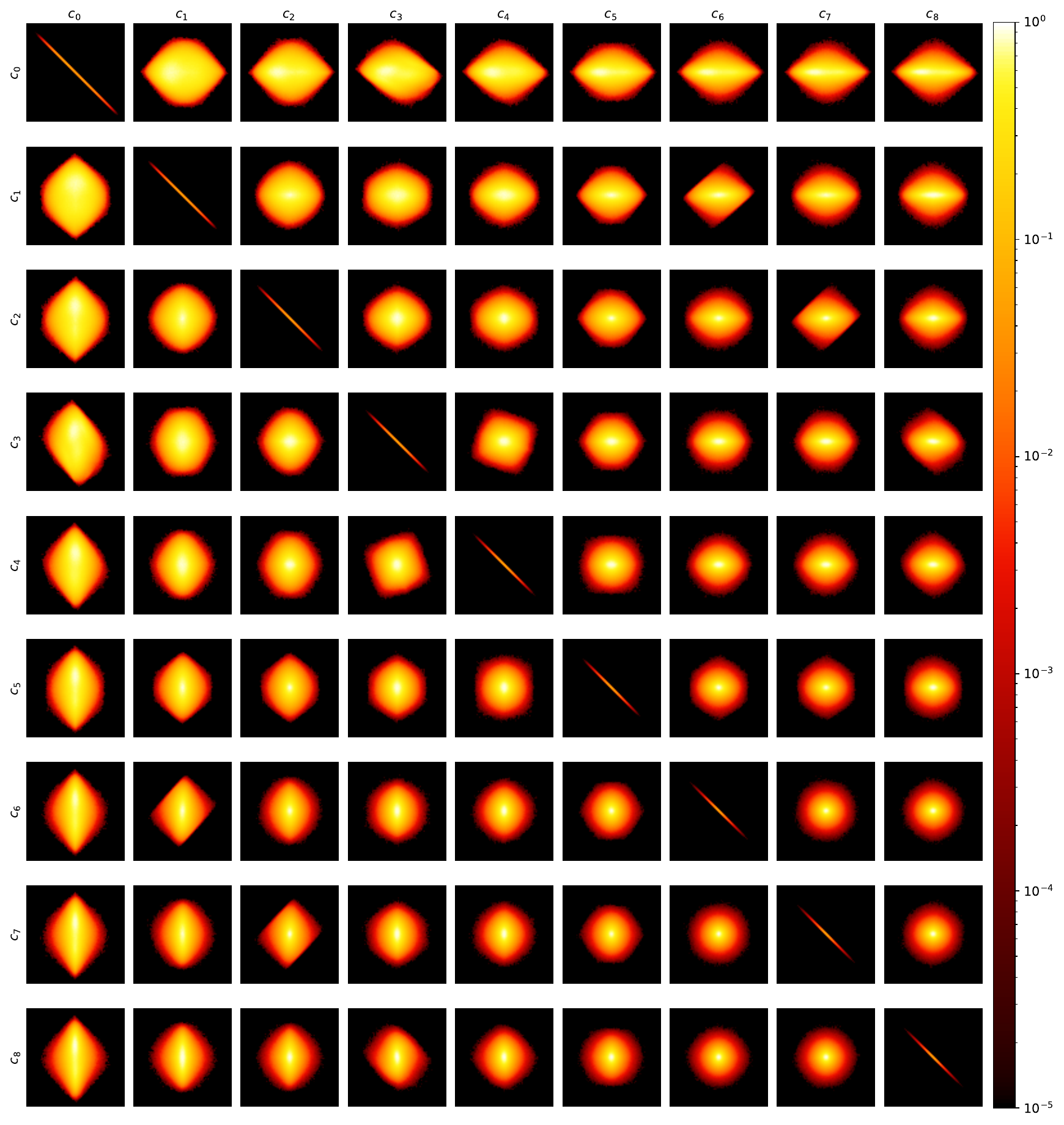}
    \caption{Bi-variate coefficient scatter plot of the phenotype \textit{sun}.}
    \label{fig:pheno_full_sun}
\end{figure*}

The main paper showed only scatter plots between two select coefficient distributions $c_i$ and $c_j$. Here we include the all bi-variate scatter plots of selected examples for each phenotype over all pairs of distributions (i.e. $i=0,\dotsc, 8$ and $j=0,\dotsc, 8$):

\cref{fig:pheno_full_all} shows the scatter plots over all filters that we have extracted;\cref{fig:pheno_full_spikes} shows \textit{spikes} of filters that belong to the visual category \textit{medical ct}; \cref{fig:pheno_full_symbols} shows \textit{symbols} based on filters that belong to an \textit{EfficientNet-l2-ns-475} pretrained on the massive \textit{JFT-300m} and fintetuned on \textit{ImageNet1k}; \cref{fig:pheno_full_point} shows \textit{point} computed on filters of our intentionally overparameterized models trained on \textit{MNIST}; and \cref{fig:pheno_full_sun} shows \textit{spikes} computed on filters of the task \textit{depth estimation}.

\section{Training of low resolution models}
\label{sec:low_res_training}

Models were taken from \cite{huy_phan_2021_4431043} and slightly modified by us to support different input channel and class modalities. Additionally, some more architectures were added. Generally, these models are quite similar to the architectures proposed in their respective original publications. However typically, Pooling will be reduced, dilated or strided convolutions will be replaced by regular convolutions, and convolution kernel sizes are reduced to be no larger than $3\times 3$.

All models are trained on NVIDIA A100 GPUs and hyper-parameters independent of the dataset. Stochastic matrix multiplication is turned off via cuDNN settings.
Inputs are scaled to $32\times 32$ px and channel-wise normalized. CIFAR data is additionally zero-padded by 4 px along each dimension, and then transformed using a $32\times 32$ random crops, and random horizontal flips. For the hyper parameters an initial learning rate of 1e-8, a weight decay of 1e-2, a batch-size of 256 and a nesterov momentum of 0.9 is used. A SGD optimizer is used, and scheduled to linearly increase the learning rate on each step for the first 30 epochs to 1e-1. Then, a cosine annealing schedule follows for the remaining 70 epochs. The loss is determined using Categorical Cross Entropy. Results are reported in \cref{tab:lowres_results}.

\section{Training of ResNet-9 variants on CIFAR-10}

The \textit{ResNet-9} models were trained as detailed in \cref{sec:low_res_training}. However, the different random seed were provided for each model. Results are reported in \cref{tab:resnet_9_retraining}.

\begin{table}
    \small
    \centering
    \caption{Performance of retrained ResNet-9 models with random seeds obtained after the validation epoch with the highest validation accuracy.}
    \label{tab:resnet_retrain_results}
    \begin{tabularx}{\linewidth}{lXXXXX}
        \toprule
        \textbf{Model ID} & \textbf{Best \newline Epoch} & \textbf{Train \newline Loss} & \textbf{Train \newline Accur.} & \textbf{Valid. \newline Loss} & \textbf{Valid. \newline Accur.} \\
        \midrule
        resnet9\_0 & 93 & 0.016 & 99.996 & 0.174 & 94.792 \\
        resnet9\_1 & 94 & 0.016 & 99.980 & 0.176 & 94.631 \\
        resnet9\_2 & 96 & 0.016 & 99.986 & 0.177 & 94.571 \\
        resnet9\_3 & 89 & 0.017 & 99.976 & 0.175 & 94.812 \\
        resnet9\_4 & 99 & 0.015 & 99.992 & 0.175 & 94.762 \\
        resnet9\_5 & 94 & 0.016 & 99.994 & 0.174 & 94.822 \\
        resnet9\_6 & 91 & 0.016 & 99.986 & 0.175 & 94.812 \\
        resnet9\_7 & 94 & 0.016 & 99.994 & 0.173 & 94.852 \\
        resnet9\_8 & 91 & 0.017 & 99.992 & 0.174 & 94.862 \\
        resnet9\_9 & 96 & 0.016 & 99.988 & 0.178 & 94.832 \\
        \bottomrule
    \end{tabularx}
    \label{tab:resnet_9_retraining}
\end{table}


\clearpage
\onecolumn
\footnotesize
%
\begin{longtable}{llrrrrr}
    \caption{Performance of low resolution models with random seeds obtained after the validation epoch with the highest validation accuracy.}\label{tab:lowres_results}\\
    \toprule
    \textbf{Dataset} & \textbf{Model} & \textbf{Best Epoch} & \textbf{Train Loss} & \textbf{Train Acc.} & \textbf{Val. Loss} & \textbf{Val. Acc.} \\
    \midrule
    \endfirsthead
    \toprule
    \textbf{Dataset} & \textbf{Model} & \textbf{Best Epoch} & \textbf{Train Loss} & \textbf{Train Acc.} & \textbf{Val. Loss} & \textbf{Val. Acc.} \\
    \midrule
    \endhead
    \midrule
    \multicolumn{7}{c}{{Continued on next page}} \\
    \midrule
    \endfoot
    
\bottomrule
\endlastfoot
cifar10 & alexnet & 95 & 0.563 & 80.717 & 0.604 & 79.297 \\
cifar10 & densenet121 & 96 & 0.047 & 99.716 & 0.247 & 93.780 \\
cifar10 & densenet161 & 97 & 0.042 & 99.830 & 0.232 & 94.311 \\
cifar10 & densenet169 & 98 & 0.046 & 99.742 & 0.235 & 94.171 \\
cifar10 & googlenet & 97 & 0.064 & 99.651 & 0.242 & 92.919 \\
cifar10 & inception\_v3 & 96 & 0.062 & 99.501 & 0.254 & 93.550 \\
cifar10 & mobilenet\_v2 & 96 & 0.074 & 98.892 & 0.237 & 93.760 \\
cifar10 & resnet101 & 96 & 0.041 & 99.329 & 0.242 & 93.399 \\
\end{longtable}

\clearpage

\begin{longtable}{p{0.12\linewidth}p{0.1\linewidth}p{0.66\columnwidth}}
\caption{Description of columns present in the meta data.}\label{tab:meta_columns}\\
\toprule
\textbf{Column} & \textbf{Values} & \textbf{Description} \\
\midrule
\endfirsthead
\toprule
\toprule
\textbf{Column} & \textbf{Values} & \textbf{Description} \\
\midrule
\endhead
\midrule
\multicolumn{3}{c}{{Continued on next page}} \\
\midrule
\endfoot

\bottomrule
\endlastfoot

model\_id & int & Unique ID of the model. \\
conv\_depth & int & Convolution depth of the extracted filter i.e. how many convolution layers were hierarchically below the layer, that this filter was extracted from. \\
conv\_depth\_norm &  float & Similar to conv\_depth but normalized by the maximum conv\_depth. Will be a float between 0 (first layers) $\dotsc$, 1 (towards head). \\
filter\_ids & list of ints & List of Filter IDs that belong to this record. These can directly be mapped to the rows of the filter array. \\
model & str & Unique string ID of the model. Typically, but not reliably in the format \texttt{\{name\}\_\{trainingset\}\_\{onnx opset\}}. \\
producer & str & Producer of the ONNX export. Typically various versions of PyTorch. \\
op\_set & int & Version of the ONNX operator set used for export. \\
depth & int & Total hierarchical depth of the model including all layers. \\
\end{longtable}

\clearpage

\renewcommand\_{\textunderscore\allowbreak}

\begin{longtable}{p{6cm}p{2cm}p{2cm}p{2cm}p{1.2cm}p{1cm}}
\caption{List of all collected models. Mainly sourced from \cite{rw2019timm,huy_phan_2021_4431043,croce2021robustbench}. Where possible, the dataset IDs correspond to the \textit{TensorFlow} naming \cite{tensorflow}. Models that start with \textit{hso\_} denote out intentionally overparametrized models (\textbf{h}ighly \textbf{s}parse and \textbf{o}verparametrized).}\label{tab:models}\\
\toprule
                                                                                                                                     \textbf{Model ID} & \textbf{Pretraining-\linebreak Dataset} &                                                                                         \textbf{Training-\linebreak Dataset(s)} &                   \textbf{Task} & \textbf{Visual\linebreak Category} &  \textbf{$3\times 3$\linebreak Filters} \\
\midrule
\endfirsthead
\toprule
                                                                                                                                     \textbf{Model ID} & \textbf{Pretraining-\linebreak Dataset} &                                                                                         \textbf{Training-\linebreak Dataset} &                   \textbf{Task} & \textbf{Visual\linebreak Category} &  \textbf{$3\times 3$\linebreak Filters} \\
\midrule
\endhead
\midrule
\multicolumn{6}{c}{{Continued on next page}} \\
\midrule
\endfoot

\bottomrule
\endlastfoot
                                                                                                agripredict\_disease\_classification\_prop\_11 \cite{agripredict} &                   - &                                                                                              agripredict &         Classification &          plants &         7232 \\
                                                                                           cyclegan\_horse2zebra\_g\_11 \cite{zhu2020unpaired} &                   - &                                                                                    cycle\_gan/horse2zebra &          GAN-Generator &         natural &      1220608 \\
                                                                                           cyclegan\_monet2photo\_g\_11 \cite{zhu2020unpaired} &                   - &                                                                                    cycle\_gan/monet2photo &          GAN-Generator &             art &      1220608 \\
                                                                                   cyclegan\_facades\_label2photo\_g\_11 \cite{zhu2020unpaired} &                   - &                                                                                        cycle\_gan/facades &          GAN-Generator &         natural &      1220608 \\
                                                                                         cyclegan\_style\_cezanne\_g\_11 \cite{zhu2020unpaired} &                   - &                                                                                  cycle\_gan/cezanne2photo &          GAN-Generator &             art &      1220608 \\
\end{longtable}
\twocolumn